\documentclass[11pt,a4paper,reqno]{amsart} 
\usepackage{amsaddr} 
\usepackage{graphicx}
\usepackage{comment}
\usepackage{mathrsfs}
\usepackage{scalerel,stackengine}
\usepackage{adjustbox}
\usepackage{xcolor}
\usepackage{braket}
\usepackage[shortlabels]{enumitem}
\usepackage{nicematrix}
\usepackage{eqparbox}
\usepackage{xfrac}
\usepackage{subcaption}
\usepackage{geometry}
\usepackage{caption}
\usepackage{comment}
\usepackage{float}
\usepackage{hyperref}
\hypersetup{hidelinks}

\newcommand{\matheqbox}[3][c]{\eqmakebox[#2][#1]{$#3$}}

\newcommand\scalemath[2]{\scalebox{#1}{\mbox{\ensuremath{\displaystyle #2}}}}

\theoremstyle{definition}
\newtheorem{defn}{Definition}[subsection]
\newtheorem{thm}[defn]{Theorem}
\newtheorem{prop}[defn]{Proposition}

\newtheorem{rem}[defn]{Remark}

\numberwithin{equation}{section}          

\title{The geometry of BERT}

\author{Matteo Bonino$^{\dag}$}
\address{Department of Mathematics "Giuseppe Peano", University of Turin, Via Verdi 8, 10124 Turin, Italy}
\email{matteo.bonino@unito.it}

\author{Giorgia Ghione$^{\dag}$}
\address{Department of Electronics and Telecommunications, Politecnico di Torino, Corso Duca degli Abruzzi 24, 10129 Turin, Italy}
\email{giorgia.ghione@polito.it}

\author{Giansalvo Cirrincione}
\address{University of Picardie Jules Verne, Chemin du Thil, 80025 Amiens, France}
\email{nimzoexin59@gmail.com}

\thanks{$^\dag$These authors contributed equally.}

\begin{document}

\begin{abstract}
Transformer neural networks, particularly Bidirectional Encoder Representations from Transformers (BERT), have shown remarkable performance across various tasks such as classification, text summarization, and question answering. However, their internal mechanisms remain mathematically obscure, highlighting the need for greater explainability and interpretability. 
In this direction, this paper investigates the internal mechanisms of BERT proposing a novel perspective on the attention mechanism of BERT from a theoretical perspective. 
The analysis encompasses both local and global network behavior. At the local level, the concept of directionality of subspace selection as well as a comprehensive study of the patterns emerging from the self-attention matrix are presented. Additionally, this work explores the semantic content of the information stream through data distribution analysis and global statistical measures including the novel concept of cone index. 
A case study on the classification of SARS-CoV-2 variants using RNA which resulted in a very high accuracy has been selected in order to observe these concepts in an application.
The insights gained from this analysis contribute to a deeper understanding of BERT’s classification process, offering potential avenues for future architectural improvements in Transformer models and further analysis in the training process.

\end{abstract}

\maketitle



\section{Introduction} 

In recent years, Transformer neural networks \cite{2017arXiv170603762V}, particularly Bidirectional Encoder Representations from Transformers (BERT) \cite{DevlinBERT}, have been extensively utilized across a wide range of tasks, including classification, text summarization, question answering, text generation, and forecasting \cite{kamath2022transformers}. Despite their remarkable performance in these diverse domains, the underlying mechanisms driving their functionality are still mathematically quite obscure.

The explainability and interpretability of Transformer models, especially BERT, have thus emerged as crucial areas of research \cite{kamath2022transformers}. These models' complexity and black-box nature drive the need for tools and techniques in order to understand their decision-making processes.  

The behavior of Transformers can be analyzed in two distinct phases: training and recall. During the training phase, the model develops its internal structures, while the recall phase leverages these structures for inference. 

This work presents a novel approach to understand the deep learning processes within the BERT architecture. The objective is to uncover the underlying structures formed during training and to gain insights into how these structures are employed during classification inference. Such an investigation is significant both for shedding light on the model's black box nature and for establishing a foundation for potential architectural advancements in the future.

This work introduces two main novel concepts. The first one is the directionality related to the subspace selection in the data processing mechanism of BERT, to assess the way it drives self-attention.
The second one is the concept of cone index in the information content of the input sequences, which is a metric to understand its internal coherence. Additionally, the patterns arising in the self-attention matrices are studied and classified. Finally, the stream of information is modelled throughout the layers of BERT, drawing conclusions on the specific role each component of BERT plays.

To facilitate this analysis, a case study with a notably high classification accuracy has been selected in order to explore how the trained architecture achieves such performance. The examination of this case aims to gain insights into the operational mechanisms that drive the model's success during the recall phase. The considered case study deals with the classification of Sars-CoV-2 variants based on the RNA sequence of the virus, presented in \cite{ghione2022interpretable}, \cite{Ghione2023}, \cite{articoloIEEEAccess}.

The work is organized as follows: Section 2 provides a review of the relevant literature, focusing on explainability and interpretability in Transformer models, particularly BERT. Section 3 describes the structure of BERT and presents the case study. In Section 4, the local analysis, i.e. the analysis of each head's behaviour, is studied. Section 5 discusses the attention patterns arising from the self-attention process, while Section 6 introduces the concept of cone index and presents the study of the stream of information. Finally, Section 7 concludes the paper, outlining key findings and suggesting directions for future research.

\section{Related work}
Below is a state-of-the-art overview of methods to explain and interpret Transformer models, including the analysis of attention patterns.

\subsection{Attention mechanisms for interpretability}
Transformers like BERT rely on attention mechanisms, which highlight the relevance of tokens in a sequence for model predictions. BERTviz is a widely used tool that visualizes attention across different heads and layers, providing insights into token interactions \cite{vig2019}. However, attention weights alone may not reliably explain model decisions, as highlighted in \cite{jain2019}, who argue that attention weights are frequently uncorrelated with token importance in final predictions.

\subsection{Attention rollout}
Attention rollout is a technique for the aggregation of attention weights across all layers to provide a more global view of how information flows through the Transformer model. 
By multiplying attention weights across layers, attention rollout traces the flow of information from the input to the output \cite{abnar2020}. This method tries to establish a global view of token influence throughout the model and can be seen as a way to quantify information flow beyond single-layer visualizations.

\subsection{Analysis of attention patterns in heads}
Attention heads in BERT have been shown to capture different types of syntactic and semantic information, leading to investigations into the patterns that emerge in their attention maps. Certain heads focus on specific syntactic relationships like subject-verb or noun-adjective dependencies, while others capture long-range dependencies between distant tokens, as shown in \cite{clark2019}. Research has shown that attention heads specialize in different linguistic roles, and some attention heads can be pruned without significantly degrading model performance, indicating redundancy in the attention mechanisms.

For instance, in \cite{voita2019} it has been found that in multi-head self-attention, certain heads are specialized in specific linguistic tasks, while others appear to contribute little to the model’s overall predictions. These findings are particularly useful for interpretability, as they show which parts of the attention mechanism are most relevant to the model’s predictions and which can be considered less critical.

\subsection{Probing classifiers}
Probing classifiers are auxiliary models trained to predict linguistic properties (e.g., syntactic dependencies, parts of speech) from the internal representations of BERT. These probes reveal the types of information encoded in different layers. Studies have shown that lower layers capture more syntactic features, while higher layers encode semantic features \cite{tenney2019}, providing insights into how BERT processes language hierarchically.

\subsection{Feature importance methods}
Feature importance techniques like LIME \cite{ribeiro2016} and SHAP \cite{lundberg2017} explain individual predictions by perturbing inputs or calculating importance scores for each token. These methods help to identify the most influential tokens for specific predictions. Another proposed method, Integrated Gradients, attributes model outputs back to input features involving the computation of the gradient of the output with respect to the input \cite{sundararajan2017}.

\subsection{Model-specific explanation methods}
Some explanation techniques are designed specifically for BERT and other Transformer models. For instance, Layer-wise Relevance Propagation (LRP) is adapted to propagate relevance through the attention heads and layers, allowing for a detailed understanding of how individual tokens contribute to the model's predictions \cite{voita2019}. Similarly, DeepSHAP combines Shapley values with deep learning-specific methods to efficiently explain deep models like BERT \cite{lundberg2017}.

\subsection{Influence functions}
Influence functions trace predictions back to the most influential training data. This technique allows researchers to identify which training examples had the most impact on a particular prediction, offering a data-centric perspective on explainability \cite{koh2017}. Influence functions help link BERT’s predictions to specific patterns in the training data, making the model's decision-making more transparent.

\subsection{Contrastive explanations}
Contrastive explanations focus on explaining why one prediction was made instead of another. In BERT, methods like counterfactual reasoning and adversarial perturbation can identify minimal changes to the input that would result in different predictions \cite{goyal2019}. These techniques provide insights into decision boundaries and critical factors affecting the model’s choices.

\subsection{Saliency maps}
Saliency maps highlight which parts of the input are most important for the model's prediction. In BERT, gradient-based saliency maps and Layer-wise Relevance Propagation (LRP) are often used to pinpoint the most relevant tokens for a given output \cite{sundararajan2017}. These maps provide visual explanations of the model’s focus during inference.

\subsection{Head and layer pruning}
Pruning techniques have been employed to explore the significance of individual attention heads and layers in BERT. By selectively removing attention heads, researchers can assess their importance to the model's performance. Studies have shown that many attention heads can be pruned without significantly impacting the model’s accuracy, suggesting redundancy in the attention mechanism \cite{voita2019}. This finding supports the idea that only a subset of heads is critical for specific tasks, making it possible to simplify the model without large performance losses.

\subsection{Evaluation of explanation methods}
Metrics for evaluating explainability methods focus on two key criteria: faithfulness (whether the explanation reflects the true reasoning process of the model) and plausibility (whether the explanation is intuitive to humans). The ERASER benchmark has been developed to standardize the evaluation of explainability methods for NLP models, ensuring that explanations are both accurate and interpretable \cite{deyoung2020}.

\section{Preliminaries}

\subsection{BERT architecture and notation}

\begin{figure}[h!]
    \centering
    {\includegraphics[width=0.4\linewidth]{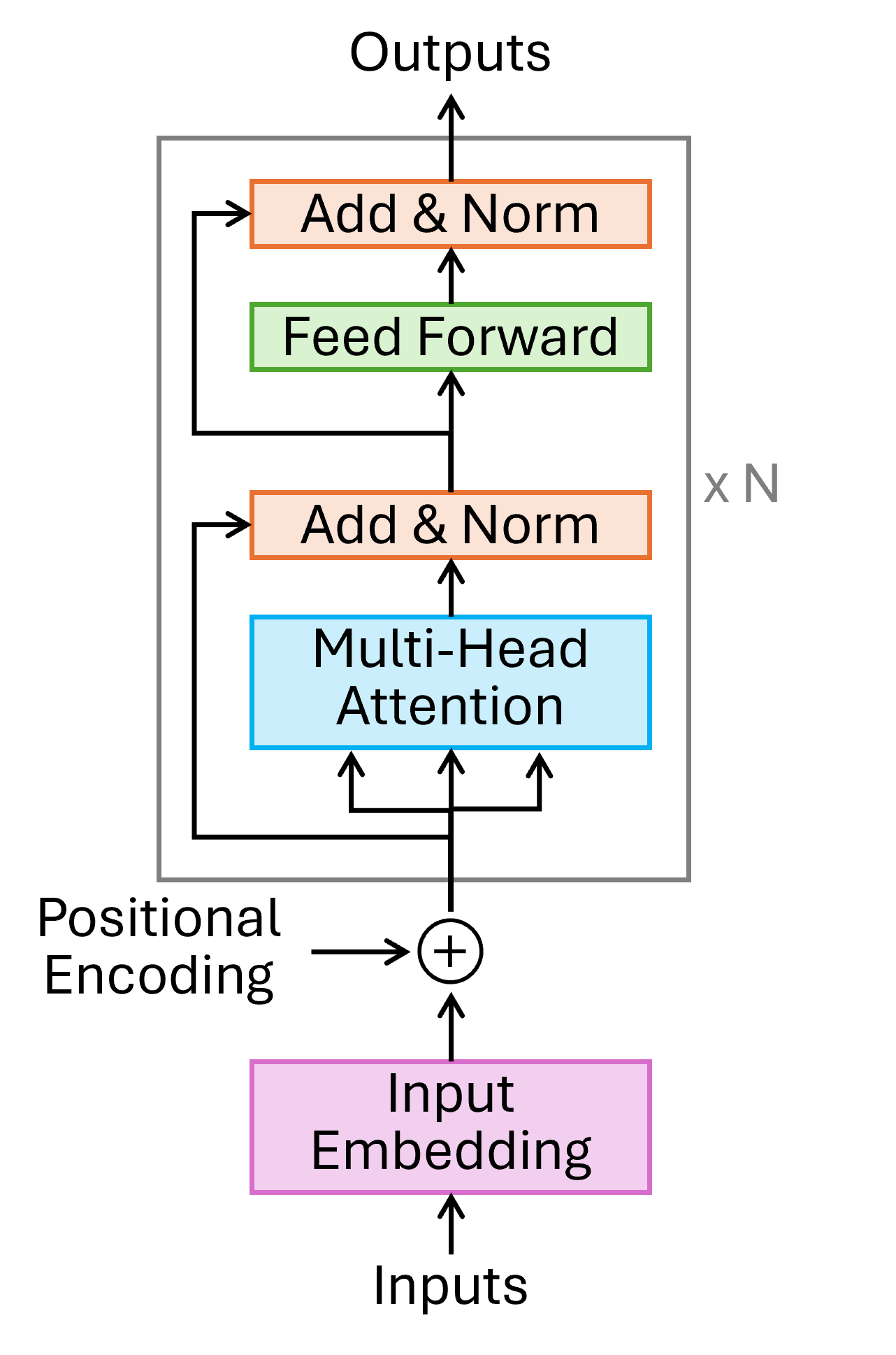}}
    \caption{BERT architecture introduced in \cite{DevlinBERT}.}
    \label{fig:bert}
\end{figure}

This work utilizes the Bidirectional Encoder Representations from Transformers (BERT), a state-of-the-art neural network based on the Transformer architecture for generating language models \cite{DevlinBERT}. BERT structure primarily consists of the encoder stack from the original Transformer model introduced in \cite{2017arXiv170603762V}, as shown in Fig. \ref{fig:bert}. It features a bidirectional encoder with $L$ layers, each comprising $A$ attention heads, which processes input sequences of up to 512 tokens and produces an embedding vector (known as the hidden state) of size $H=d_{model}$ for each position, matching the size of the input embeddings. This output can serve as the input for various tasks, including classification. The original BERT was proposed in two sizes: 1) $BERT_{BASE}$ with $L=12, A=12, H=768$, totalling 110 million parameters, and 2) $BERT_{LARGE}$ with $L=24, A=16, H=1024$, totalling 340 million parameters.

The initial step in BERT involves generating embeddings from input sequences. Initially, these sequences are tokenized and converted into numerical form by mapping each token to a unique integer identifier in the vocabulary being considered. Subsequently, each token is associated with a $d_{model}$-dimensional embedding vector, which is optimized during training. These resulting vectors are then stacked to create a matrix of dimensions $n \times d_{model}$, where $n$ represents the number of tokens in the sequence.
Input embeddings are adjusted to encode information regarding the token order in the sequence. This step is crucial because the Transformer network processes each entire sequence in parallel without recurrence or correlation mechanisms that would otherwise retain the relative and absolute token positions. To address this issue, the approach proposed in \cite{2017arXiv170603762V} involves incorporating predefined positional encodings of size $n \times d_{model}$ using sine and cosine functions without the need to train them:

\begin{equation}
PE_{(pos,2i)} = sin(pos/10000^{2i/d_{model}})
\end{equation}
\begin{equation}
PE_{(pos,2i+1)} = cos(pos/10000^{2i/d_{model}})
\end{equation}

Here, $pos$ denotes the token position and $i$ represents the index within the embedding dimension.

Then, the encoder stack of the Transformer, which is part of the BERT architecture, includes $L$ identical encoder blocks, each with unique weights, chained together. Each encoder block consists of two sub-layers: the multi-head attention layer and the fully connected layer. After processing through each sub-layer, dropout regularization \cite{JMLR:v15:srivastava14a} is applied. A residual connection is then added to mitigate the vanishing gradient problem, and the result is normalized row-wise (token-wise) using layer normalization to enhance network stability.

\subsection{Study case} 
In this study, a case study with notably high classification accuracy was chosen to perform the analysis. The selected case study involves the classification of Sars-CoV-2 variants using the RNA sequence of the virus, as presented in \cite{ghione2022interpretable}, \cite{Ghione2023}, \cite{articoloIEEEAccess}.
Since the initial emergence of the COVID-19 pandemic, the SARS-CoV-2 virus captivated the attention of scientific communities globally. This intense focus has resulted in significant insights into the biological structure of the virus, its modes of transmission, and strategies for mitigating its spread. A critical and pressing issue was the ongoing need for the detection and monitoring of circulating SARS-CoV-2 variants. Specifically, medical laboratories worldwide were faced with the challenge of accurately and reliably classifying these variants based on their genomic sequences. In order to facilitate the study of the virus and its variants, the publicly accessible data-sharing platform GISAID \cite{gisaid} collected and made available more than 13 million sequences belonging to multiple SARS-CoV-2 variants from all over the globe, such as  Alpha (B.1.1.7+Q.x), Beta (B.1.351+B.1.351.2+.1.351.3), Gamma (P.1+P.1.x), Delta (B.1.617.2+AY.x), Omicron (B.1.1.529+BA.*), GH/490R (B.1.640+B.1.640.*), Lambda (C.37+C.37.1), and Mu (B.1.621 + B.1.621.1). For the purpose of this work, the GISAID SARS-CoV-2 repository was accessed in August 2022 and a total of 49403 distinct sequences were collected for classes Alpha, Beta, Gamma, Delta, Omicron, Gh, Lambda and Mu. Each RNA sequence was aligned to the reference sequence \emph{SARS-CoV-2 isolate Wuhan-Hu-1} \cite{refSequence} and the S gene was extracted, due to its crucial role in COVID-19 infection: indeed, this gene encodes the spike glycoprotein, which is essential for the virus entry into host cells \cite{xia2021domains}.  
The resulting dataset was divided into a training set consisting of 39,522 sequences and a test set comprising 9,881 sequences, following a stratified 80\%-20\% split.
Then, the sequences were segmented into 12-base long k-mers (tokens) with a stride of 9 bases, and unique k-mers were incorporated into the BERT vocabulary.

The chosen model for variant classification is $BERT_{BASE}$, which can be found in the HuggingFace library \cite{huggingface}. The model is composed of $L=12$ layers with $A=12$ heads each, the hidden size $H$ was set to 768, and the dimensionality of the feed-forward layer to 3072. GeLU activation \cite{DBLP:journals/corr/HendrycksG16}, absolute positional encoding and a dropout probability equal to 0.1 were used. The fine-tuning of the model was performed via mini-batches of 4 sequences for 2 epochs with cross-entropy loss, AdamW optimizer \cite{DBLP:journals/corr/abs-1711-05101}, and a learning rate warm-up schedule. To enable the classification task of this work, a special classification token (CLS) was inserted at the beginning of each sequence and a classification head was added on top of $BERT_{BASE}$. This head consists of an MLP with a hidden layer of 768 units that takes the CLS token embeddings output by the encoder as input (with dropout and Tanh activation), an output layer with $C$ units (where $C$ is the number of classes), and softmax layer to produce the classification result.

\section{Local analysis} 

This section is devoted to investigating the role attention plays locally, that is at every head level. It consists of two subsections: the first one focuses on the mechanism of attention searching, while the second one discusses rank collapse.

\subsection{Self-attention mechanism}

All the neural networking involved in the classification is deeply related to the attention seeking process, consisting of the matrix $Q$ questioning the matrix $K$ by means of a sort of non-commutative scalar product. \\
BERT's input $X \in \mathbb{R}^{512 \times 768}$ is mapped into the matrices $Q$, $K$ and $V$ by means of the affine transformations
\begin{equation}
\begin{cases}
Q= X W_Q + \mathbf{1} b_Q^T,\\
K= X W_K + \mathbf{1}b_K^T,\\
V= X W_V + \mathbf{1}b_V^T.\\
\end{cases}
\end{equation}
Here $W_{Q,K,V}$ are rectangular matrices of size $768 \times 64$, while $b_{Q,K,V}$ are bias vectors of size $512 \times 64$ which play the role of corrective factors and $\mathbf{1}$ is the all-one column vector. In this step, the input is investigated in order to highlight some features and whiten others less effective. These matrices, especially $W_Q$ and $W_K$, work as \textbf{searching engines} which scan the embedded sequences and detect the mutual information links.

\begin{figure*}[h!]
    \centering
    {\includegraphics[width=\linewidth]{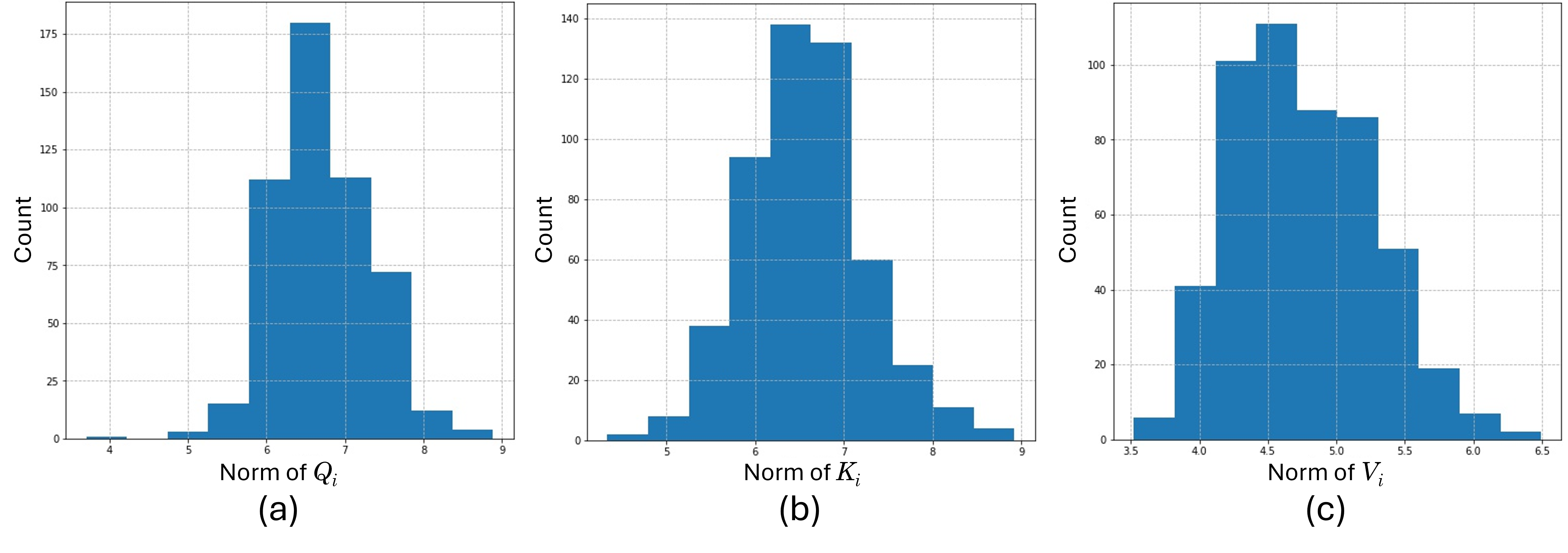}}
    \caption{Histograms of the norms of the row-vectors $Q_i, K_i$ and $V_i$ of the matrices $Q,K$ and $V$, respectively, of a correctly classified sequence in head 4, layer 1.}
    \label{fig:norm_1_4}
\end{figure*}

Figure \ref{fig:norm_1_4} shows an example of histogram of the row-wise norms of the matrices $Q, K$ and $V$ from head 4, layer 1 calculated using a correctly classified sequence,  whose corresponding maximal singular values are $mSV_{W_Q}=1.72, mSV_{W_K}=2.68$, and $mSV_{W_V}=0.98$. Notice that there can be found a frequent amplification and a maximal singular value usually larger than 1, which would indicate the most relevant stretching direction (particularly in the case of the matrices $Q$ and $K$ which contribute to the self-attention process). 

As a matter of fact, each head draws features-driven directions that convey the major found information at that level. This process outlines a \textbf{subspace selection} establishing a hierarchy between the possible directions in the immersion space $\mathbb R^{512 \times 768}$. It is hence possible to define the \textbf{direction cone} of the engine process as the collection of those vectors facing the highest stretching.

The final directional selection of the self-attention mechanism follows from the resulting coherence between $Q$ and $K$ computed in the matrix $S$
\begin{equation}
S=\mathrm{softmax} \left( \frac{Q K^T}{\sqrt{d_k}} \right)
\end{equation}
where the row-wise softmax function task is precisely to show the selected searching direction whitening the background noise. Here, $d_k$ is the correcting term related to dimensionality, which is $64$ in the case study, while the row-wise softmax function applied to a matrix $Y \in \mathbb{R}^{n \times k}$ is defined as
\begin{equation}
\mathrm{softmax}(Y)=\begin{pNiceMatrix}
\frac{e^{(Y_1)_1}}{\sum_{j=1}^k e^{(Y_1)_j}} & \frac{e^{(Y_1)_2}}{\sum_{j=1}^k e^{(Y_1)_j}} & \Cdots &\frac{e^{(Y_1)_k}}{\sum_{j=1}^k e^{(Y_1)_j}}\\
\frac{e^{(Y_2)_1}}{\sum_{j=1}^k e^{(Y_2)_j}} & \frac{e^{(Y_2)_2}}{\sum_{j=1}^k e^{(Y_2)_j}} & \Cdots &\frac{e^{(Y_2)_k}}{\sum_{j=1}^k e^{(Y_2)_j}}\\
\Vdots & \Vdots & \Ddots & \Vdots \\
\frac{e^{(Y_n)_1}}{\sum_{j=1}^k e^{(Y_n)_j}} & \frac{e^{(Y_n)_2}}{\sum_{j=1}^k e^{(Y_n)_j}} & \Cdots& \frac{e^{(Y_n)_k}}{\sum_{j=1}^k e^{(Y_n)_j}}\\
\end{pNiceMatrix}
\end{equation}
\noindent the $Y_i$ being the rows of $Y$. Notice that the product $QK^T$ can be written as
\begin{equation}
\begin{aligned}
QK^T & =(X W_Q +\mathbf{1} b_Q^T)(X W_K + \mathbf{1} b_K^T)^T\\
&=X W_Q W_K^T X^T  + \mathbf{1}b_Q^T W_K^T X^T + X W_Q b_K \mathbf{1}^T+ \mathbf{1} b_Q^T b_K \mathbf{1}^T\\
&=X M X^T + \mathbf{1}b_Q^T W_K^T X^T + X W_Q b_K \mathbf{1}^T+ \mathbf{1} b_Q^T b_K \mathbf{1}^T
\end{aligned}
\end{equation}
where 4 different terms can be distinguished. It is immediate to check that the latter two are bias adders which act row-wise. Therefore, in light of the property valid for any vector $\mathbf{x}$ and any constant $c$ \cite{mh}
$$
\mathrm{softmax}(\mathbf{x} +c)=\mathrm{softmax}(\mathbf{x}) 
$$
their contribution is null. It remains a bias correlated correcting term $\mathbf{1}b_Q^T W_K^T X^T$, and the crucial term $X M X^T$, with $M=W_Q W_K^T$, which conveys the contextual information. Notice that $M$ is not a positive semidefinite matrix; therefore it can not be interpreted as a scalar-product matrix, but is rather a \textbf{similarity measure matrix}, whose $W_Q$ and $W_K^T$ correspond to its low-rank decomposition. Unlike the cosine similarity measure, this measure is influenced by both the norms of the row-vectors (norm effect) and their relative angles (direction effect). The key role of this similarity checking (a sort of inner product), gives rise to a subspace selection related to the $S$ matrix pattern analysis. Notice that, since $M$ is not a metric matrix, there occurs an asymmetry in the self-attention mechanism. 

\subsection{Rank collapse}

Paper \cite{attnot} highlights a key issue concerning BERT's attention seeking, that is the asymptotic behaviour of rank collapse.
It is indeed a matter of fact that the mere application of the multi-head attention mechanism aims at aligning the value row-vectors in $V$, so that the rank of the output matrix tends to collapse to 1. In light of this behaviour, the presence of one perceptron after the multi-head attention block is necessary in order to counterbalance the rank loss. Nevertheless, the situation is presented as a \textit{tug of war} between the perceptron and the multi-head attention, where the perceptron's non-linearity plays the only positive role in stopping the rank loss. The new perspective presented in this paper reinterprets the rank collapse in a completely different light, showing that it may convey information content. 

In each Transformer layer, between the addition of the multi-head attention output to its input and the layer output, the embedding row-vectors are resized by means of the layer normalization (LN), by mapping every row $\mathbf{y}$ into
\begin{equation}
\mathbf{y} \mapsto \frac{\mathbf{y} - \mathbb{E}[\mathbf{y}]}{\sigma_\mathbf{y}}
\end{equation}
where the mean function $\mathbb{E}$ acts component-wise.

Here it is proven that the layer normalization does not play any role in affecting the matrix rank (see also \cite{attnot}). Indeed, notice that the following theorem holds true:
\begin{thm}
\cite{bernstein2018} Let $\mathscr{C}_1$ and $\mathscr{C}_2$ be the column spaces of two matrices $A$ and $B$ and $\mathscr{R}_1$ and $\mathscr{R}_2$ the corresponding row spaces. Then, if $A,B \in \mathbb{R}^{m \times n}$ (or $\mathbb{C}^{m \times n})$ the rank satisfies the subadditivity property 
\begin{equation}
\mathrm{rank}(A+B) \leq \mathrm{rank}(A) + \mathrm{rank}(B),
\end{equation}
\noindent where the equality holds true if and only if
\begin{equation}
\mathrm{dim}(\mathscr{C}_1 \cap \mathscr{C}_2) = \mathrm{dim}(\mathscr{R}_1 \cap \mathscr{R}_2)=0.
\end{equation}
\end{thm}

It is possible to employ the theorem in this specific case. Denote by $\mathrm{LN}\left(S\left(X\right)\right)$ the output of the sequential application of the self-attention process ($S$) and the layer normalization (LN), Also, by $\tilde{X}_{MH}$ the result of the multi-head concatenation, by $D_{LN}^{-1}$ the inverse of the diagonal matrix $512 \times 512$ of the row-wise standard deviations, and by $\mathbb{E}(\tilde{X}_{MH})$ the column vector of the row-wise means of $\tilde{X}_{MH}$. Explicitly,
\begin{equation}
\begin{aligned}
\mathrm{LN}\left(S\left(X\right)\right) &=\mathrm{LN}\left(\mathrm{Conc}\left[(S_1 V_1), (S_2 V_2 ),  \dots  (S_{12} V_{12}) \right] \right)\\
&=D_{LN}^{-1}\left( \tilde{X}_{MH} - \mathbf{1} \mathbb{E}(\tilde{X}_{MH})^T \right)
\end{aligned} 
\end{equation}
where ``$\mathrm{Conc}$" denotes the concatenation operator. It is clear that the left product with a diagonal matrix of maximal rank does not affect the rank. Then, employ the theorem choosing $A=\tilde{X}_{MH}$ and $B=\mathbf{1} \mathbb{E}(\tilde{X}_{MH})^T$, noticing that $\mathrm{rank}(\mathbf{1} \mathbb{E}(\tilde{X}_{MH})^T)=1$. Since by construction $\mathscr{C}_2= \mathscr{L}\left(\mathbb{E}(\tilde{X}_{MH}) \right)$ and by definition of the mean $\mathbb{E}(\tilde{X}_{MH}) \in \mathscr{C}_1$, it follows that
\begin{equation}
\mathrm{rank}\left(\tilde{X}_{MH} - \mathbf{1} \mathbb{E}(\tilde{X}_{MH})^T\right) < \mathrm{rank}\left(\tilde{X}_{MH} \right) +1.
\end{equation}
In order to obtain the other desired inequality, the following well-known result is needed.
\begin{prop} \cite{bernstein2018}
For any $A \in \mathbb{R}^{m \times n}$ (or $\mathbb{C}^{m \times n})$ and $\mathbf{x,y} \in \mathbb{R}^{n \times 1}$ (or $\mathbb{C}^{n \times 1})$ the following inequality is satisfied
\begin{equation}
 \mathrm{rank}\left(A + \mathbf{x}\mathbf{y}^T \right) \geq  \mathrm{rank}(A)-1,
\end{equation}
where the equality holds true if and only if
\begin{equation}
\mathbf{x} \in \mathscr{C}(A) \text{  and  } \mathbf{y} \in \mathscr{R}(A). 
\end{equation}
\end{prop}
Employing the Proposition by picking $A=\tilde{X}_{MH}$, $\mathbf{x}=\mathbf{1}$ and $\mathbf{y}^T= \mathbb{E}(\tilde{X}_{MH})^T$, it follows that
\begin{equation}
\mathrm{rank}\left(\tilde{X}_{MH} - \mathbf{1} \mathbb{E}(\tilde{X}_{MH})^T\right) > \mathrm{rank}\left(\tilde{X}_{MH} \right) -1.
\end{equation}
The combination of the two inequalities proves the independence of the layer normalization function from the rank loss.

\section{Self-attention matrix typical patterns: gates}

In the sequel, the possible different kinds of patterns arising at self-attention head level are studied by means of a qualitative and data-driven model which allows predictions and further considerations.

The mathematical model concerning the typical patterns arising in the $S$ matrix, whose first attempt at classification was established in \cite{rev}, is based on the following rules:
\begin{enumerate}
\item The self-attention matrix entries contributing to the pattern take the largest possible values, that is, if there is only one element in a row, then it is 1. 
\item The self-attention matrix entries which do not contribute to the pattern take the smallest possible values, that is, if there exists an element in the row taking part in the pattern, then the others are all 0. In such a way the background noise is whitened.
\item Both the sets of elements contributing and not contributing to the pattern have an attention score uniformly distributed, that is, if there is more than one pattern entry in a row, then they are all $\sfrac{1}{k}$, where $k$ denotes the number of entries involved. On the other hand, the non-contributing rows have scoring values all $\sfrac{1}{n}$.
\end{enumerate}

\subsection{Directional gate}

First, consider the case of the self-attention matrix showing a \textbf{vertical line pattern}. It is modelled as
\begin{equation}
S_V^{(i)}=\begin{pNiceMatrix}[
  first-row
]
 &   &  &  &{\begin{array}{c} i \\ \downarrow \end{array}} & & &  \\
 0 & 0 & \Cdots & 0 & \mathbf{1} & 0 & \Cdots & 0\\
 0 & 0 & \Cdots & 0 & \mathbf{1} & 0 & \Cdots & 0\\
\Vdots & \Vdots & & \Vdots & \Vdots & \Vdots &  & \Vdots\\
0 & 0 & \Cdots & 0 & \mathbf{1} & 0 & \Cdots & 0\\
\end{pNiceMatrix}
\end{equation}
where the subscript $V$ stands for \textit{Vertical}, while the superscript $(i)$ remarks the dependence on the column index $i$. The corresponding output matrix is
\begin{equation}
S_V^{(i)} V=\begin{pNiceMatrix}
(v^i)^T\\
(v^i)^T\\
\Vdots \\
(v^i)^T\\
\end{pNiceMatrix}
\end{equation}
where $(v^i)^T$ stands for the $i^{th}$ row in the value-matrix $V$. This is a rank-1 matrix, but this fact by itself does not exhaust the whole information transferred. In fact, the repetition of the $i^{th}$ row establishes a highly directional signal whereby for each token $j$, its attention points to the same token $i$. In light of this property, this situation has been classified as \textbf{directional gate}.

Notice that, in this situation, the rank collapse to 1 still conveys a global characteristical behaviour, since the output cone of the value-vectors has degenerated in one line with its own precise direction. Moreover, there exist cases largely studied in literature where this pattern plays a crucial role (see, for example, \cite{bertbio} and \cite{rev}). That's why the rank loss here is interpreted as anything but negative.

\begin{figure}[h!]
    \centering
    {\includegraphics[width=0.8\linewidth]{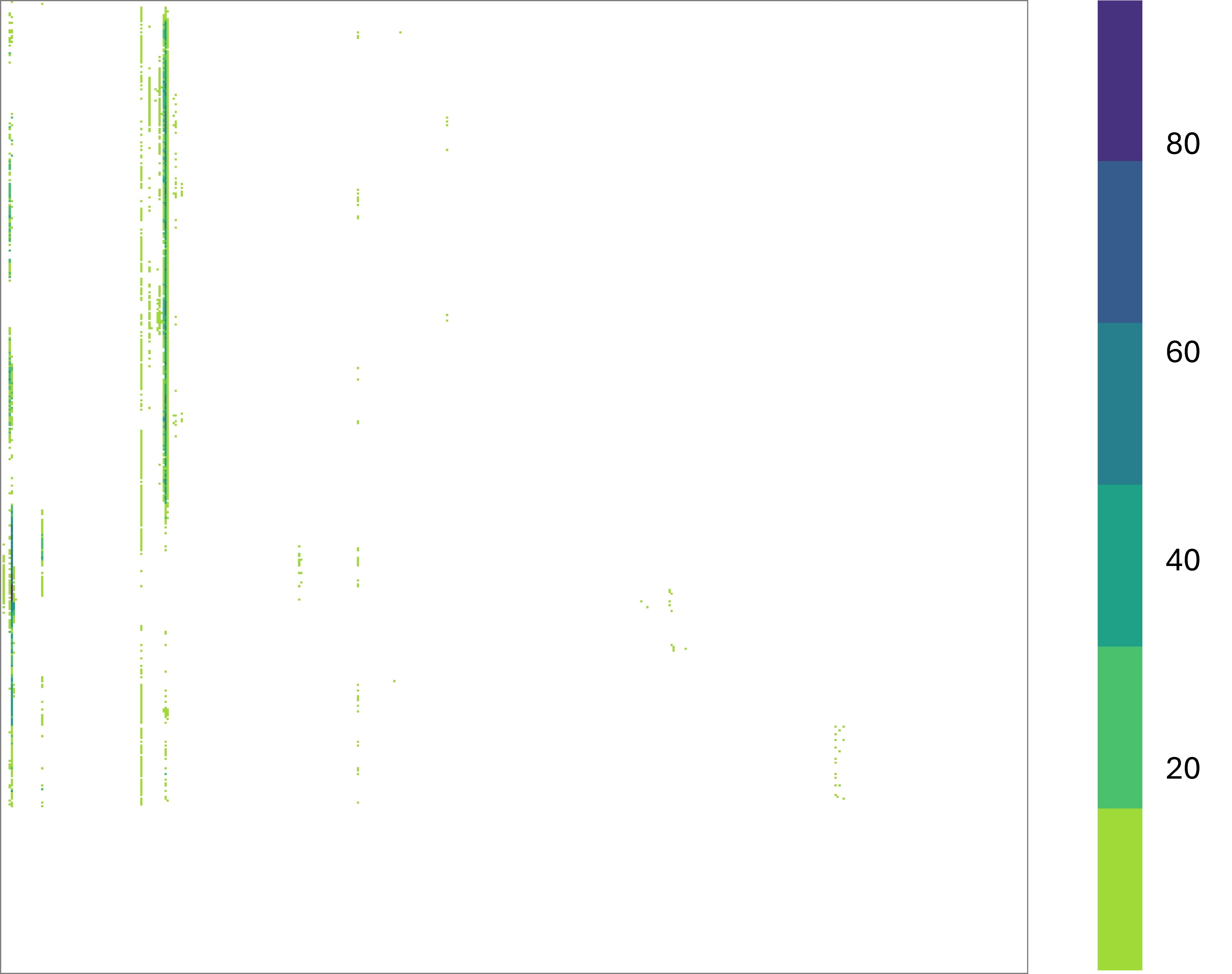}}
    \caption{Attention map of head 9, layer 4 of a correctly classified sequence, showing a directional gate.}
    \label{fig:omicron_4_9_vert}
\end{figure}

Finally, it is worth highlighting that such a pattern usually arises in relatively small blocks of vertical lines, a circumstance which witnesses that the global directionality information faces a fragmentation, which, in a certain way, localizes the phenomenon weighing the tokens' attention that contributes to the mean. In Fig. \ref{fig:omicron_4_9_vert}, an example of a directional gate taken from head 9 of layer 4 of an Omicron classified sequence is reported.

A vertical-line pattern may involve a useful rank collapse or a directional meaning towards a classifying token.

\subsection{Open gate}
Now, consider the case of the self-attention matrix showing a \textbf{diagonal pattern}. It is modelled as
\begin{equation}
S_D=\begin{pNiceMatrix}
\mathbf{1}  & 0 & 0 & \Cdots & 0 \\
 0 & \mathbf{1}  & 0 & \Cdots & 0 \\
0 & 0 & \mathbf{1}  & \Cdots & 0 \\
\Vdots & \Vdots & \Vdots & \Ddots  & \Vdots \\
0 & 0 & 0 & \Cdots & \mathbf{1}  \\
\end{pNiceMatrix}
\end{equation}
where the subscript $D$ stands for \textit{Diagonal}. Since $S_D$ equals the identity matrix, the corresponding output matrix is trivially the matrix $V$:
\begin{equation}
S_D V=V=\begin{pNiceMatrix}
(v^1)^T\\
(v^2)^T\\
\Vdots \\
(v^n)^T\\
\end{pNiceMatrix}
\end{equation}
Such a result represents the pure self-attention phenomenon. Notice that, since the matrix $S_D$ has maximum rank, it does not affect the rank of the output and neither changes the entropy. For this reason, it behaves as an \textbf{open gate}, in the sense that all the self information is carried on.

Usually, it occurs an interrupted diagonal pattern at the level of the last block of tokens (whose order of magnitude involves a dozen tokens).

\begin{figure}[h!]
    \centering
    {\includegraphics[width=0.8\linewidth]{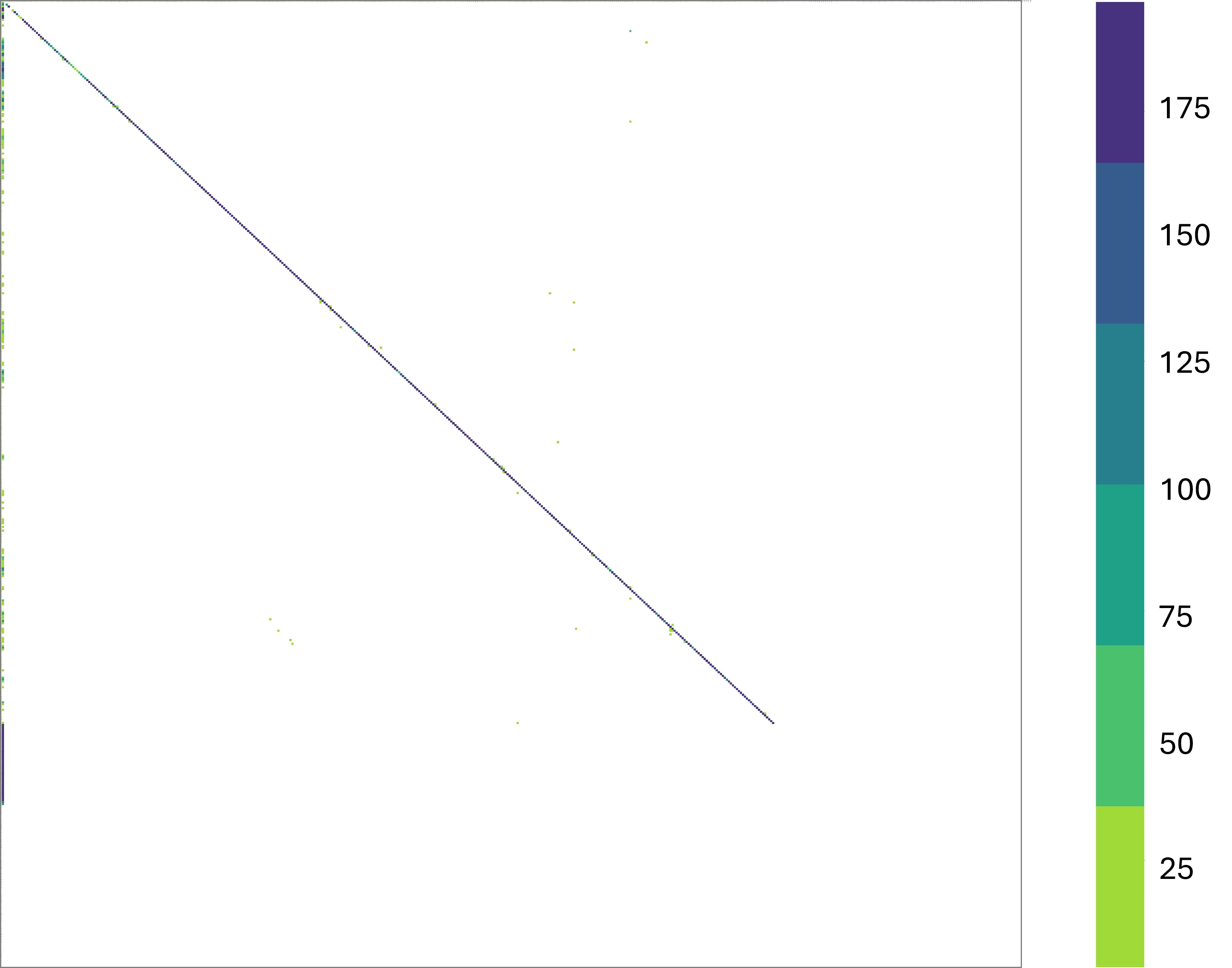}}
    \caption{Attention map of head 4, layer 2 of a correctly classified sequence, showing an open gate.}
    \label{fig:omicron_2_4_diag}
\end{figure}

In Fig. \ref{fig:omicron_2_4_diag}, an example of an open gate taken from head 4 of layer 2 of an Omicron classified sequence is reported.

\subsection{Backward contextual gate}
Take now into consideration the case of the self-attention matrix showing a \textbf{sub-diagonal pattern}. It is modelled as
\begin{equation}
S_{sD}^{(j)}=\begin{pNiceMatrix}[first-row, last-col=8
]
 &  &  &  & {\begin{array}{c} n-j \\ \downarrow \end{array}} &  & \\
\sfrac{1}{n} & \sfrac{1}{n}& \sfrac{1}{n} & \Cdots & \sfrac{1}{n} & \Cdots & \sfrac{1}{n}\\
 \Vdots & \Vdots & \Vdots &  & \Vdots &  & \Vdots\\
 \sfrac{1}{n} & \sfrac{1}{n} & \sfrac{1}{n} & \Cdots & \sfrac{1}{n} & \Cdots & \sfrac{1}{n}\\
  \mathbf{1}  & 0 & 0 & \Cdots & 0 & \Cdots & 0 &\matheqbox{a}{\leftarrow j}\\
 0 & \mathbf{1}  & 0 & \Cdots & 0 & \Cdots & 0\\
 0 & 0 & \mathbf{1}  & \Cdots & 0 & \Cdots & 0 \\
 \Vdots & \Vdots & \Vdots & \Ddots  & \Vdots & & \Vdots\\
 0 & 0 & 0 & \Cdots & \mathbf{1}  & \Cdots & 0\\
\end{pNiceMatrix}
\end{equation}
where the subscript $sD$ stands for \textit{sub-Diagonal}, while the superscript $(j)$ remarks the dependence on the row index $j$. The corresponding output matrix is
\begin{equation}
S_{sD}^{(j)} V=\begin{pNiceMatrix}[last-col=2]
\mathbb{E}\left[(v^1)^T, \cdots, (v^n)^T\right]\\
\mathbb{E}\left[(v^1)^T, \cdots, (v^n)^T\right]\\
\Vdots \\
\mathbb{E}\left[(v^1)^T, \cdots, (v^n)^T\right]\\
(v^1)^T &\matheqbox{a}{\leftarrow j}\\
(v^2)^T\\
\Vdots \\
(v^{n-(j-1)})^T\\
\end{pNiceMatrix}
\end{equation}
where $\mathbb{E}\left[(v^1)^T, \cdots, (v^n)^T\right]$ denotes the mean vector of the rows $(v^1)^T, \cdots, (v^n)^T$ of $V$. Here, the first $j-1$ rows are all equal, therefore the rank of the matrix $V$ has strictly decreased. Moreover, these rows convey just a global mean value, so that only $n-(j-1)$ rows effectively carry on processed information. The output selects just a band of values of $V$, shifted by $j$ in the sense that attention points backwards by $j$ position. This is the reason why such a pattern is called \textbf{backward contextual gate}. 

Notice that the involved information content concerns not only the rank, that is the value of the index $j$, but also the length of the sub-diagonal pattern, which usually is shorter than $n-(j-1)$. This involves a sort of \textbf{locality information} provided by a \textbf{selective contextual attention head}, which plays a crucial role in the global architecture of the Transformer classification task.

\subsection{Forward contextual gate}
Consider now the case of the self-attention matrix showing a \textbf{super-diagonal pattern}. It is modelled as
\begin{equation}
S_{SD}^{(j)}=\begin{pNiceMatrix}[first-row, last-col=9
]
 &  &  &   {\begin{array}{c} j \\ \downarrow \end{array}} & &  & & \\

0 & \Cdots & 0 & \mathbf{1}  & 0 & 0 & \Cdots & 0 \\
0 & \Cdots & 0 & 0 & \mathbf{1}  & 0 & \Cdots & 0 \\
0 & \Cdots & 0 & 0 & 0 & \mathbf{1}  & \Cdots & 0  \\
 \Vdots & & \Vdots & \Vdots & \Vdots & & \Ddots  &\Vdots\\
0 & \Cdots & 0 & 0 & 0 & 0 & \Cdots & \mathbf{1}  &\matheqbox{a}{\leftarrow j}\\
\sfrac{1}{n} & \Cdots & \sfrac{1}{n}& \sfrac{1}{n} & \sfrac{1}{n} & \sfrac{1}{n} & \Cdots & \sfrac{1}{n}\\
 \Vdots & & \Vdots & \Vdots & \Vdots & \Vdots &  & \Vdots\\
\sfrac{1}{n} & \Cdots & \sfrac{1}{n}& \sfrac{1}{n} & \sfrac{1}{n} & \sfrac{1}{n} & \Cdots & \sfrac{1}{n}\\
\end{pNiceMatrix}
\end{equation}
where the subscript $SD$ stands for \textit{Super-Diagonal}, while the superscript $(j)$ remarks the dependence on the column index $j$. The corresponding output matrix is
\begin{equation}
S_{SD}^{(j)} V=\begin{pNiceMatrix}
(v^j)^T \\
(v^{j+1})^T\\
\Vdots \\
(v^{n})^T\\
\mathbb{E}\left[(v^1)^T, \cdots, (v^n)^T\right]\\
\mathbb{E}\left[(v^1)^T, \cdots, (v^n)^T\right]\\
\Vdots \\
\mathbb{E}\left[(v^1)^T, \cdots, (v^n)^T\right]\\
\end{pNiceMatrix}
\end{equation}
where the notation is consistent with the analogous previous case. The only difference with the backward contextual gate consists in the attention being shifted forward rather than backward, since all the considerations above hold true.

\begin{figure}[h!]
    \centering
    {\includegraphics[width=0.8\linewidth]{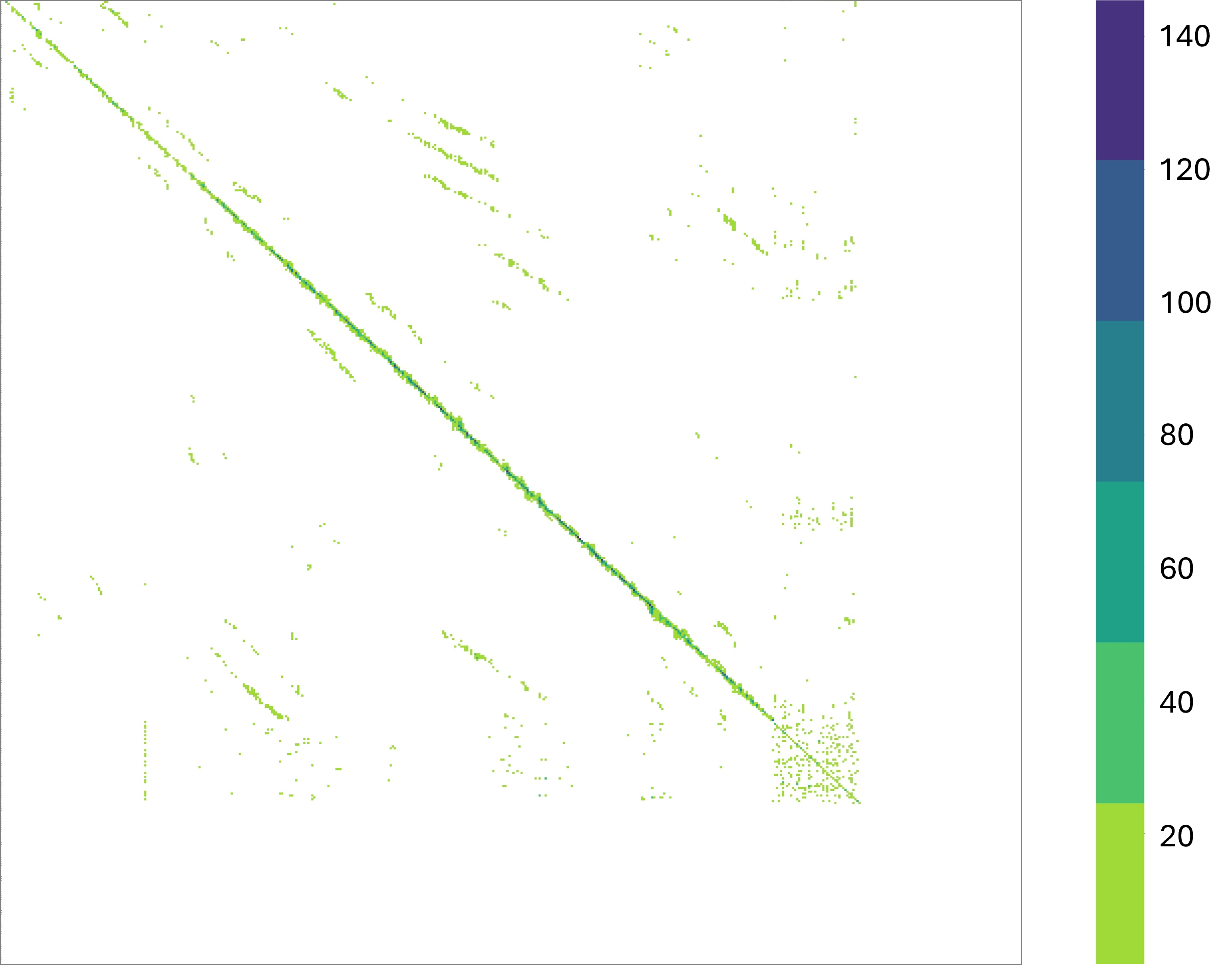}}
    \caption{Attention map of head 10, layer 7 of a correctly classified sequence, showing forward contextual gates and backward contextual gates.}
    \label{fig:omicron_7_10_contextual}
\end{figure}

Fig. \ref{fig:omicron_7_10_contextual}, an example of a pattern showing both forward contextual gates and backward contextual gates is reported, again taken by an Omicron classified sequence.

Notice the presence of a self-attention pattern: this is the usual case since sub and super diagonals usually do not occur alone.

\subsection{Cluster head}
Take into consideration the case of the self-attention matrix showing a \textbf{$k \times k$ block pattern}. It is modelled as

\begin{equation}
    \scalemath{0.8}{S_B^{(i,j;k)}=\begin{pNiceMatrix}[first-row, last-col=11
]
 &  &  &   {\begin{array}{c} i \\ \downarrow \end{array}} & &  & {\begin{array}{c} i+k \\ \downarrow \end{array}} & & & \\
\sfrac{1}{n} & \Cdots & \sfrac{1}{n}& \sfrac{1}{n} & \sfrac{1}{n} & \Cdots & \sfrac{1}{n} & \sfrac{1}{n} & \Cdots & \sfrac{1}{n}\\
\Vdots & & \Vdots & \Vdots & \Vdots & & \Vdots  &\Vdots & & \Vdots \\
\sfrac{1}{n} & \Cdots & \sfrac{1}{n}& \sfrac{1}{n} & \sfrac{1}{n} & \Cdots & \sfrac{1}{n} & \sfrac{1}{n} & \Cdots & \sfrac{1}{n}\\
0 & \Cdots & 0 & \mathbf{\sfrac{1}{k}} & \mathbf{\sfrac{1}{k}} & \Cdots & \mathbf{\sfrac{1}{k}} & 0 & \Cdots & 0 & \matheqbox{a}{\leftarrow j}\\
0 & \Cdots & 0 & \mathbf{\sfrac{1}{k}} & \mathbf{\sfrac{1}{k}} & \Cdots & \mathbf{\sfrac{1}{k}} & 0 & \Cdots & 0\\
 \Vdots & & \Vdots & \Vdots & \Vdots & & \Vdots  &\Vdots & & \Vdots \\
0 & \Cdots & 0 & \mathbf{\sfrac{1}{k}} &\mathbf{\sfrac{1}{k}} & \Cdots & \mathbf{\sfrac{1}{k}} & 0 & \Cdots & 0 & \matheqbox{a}{\leftarrow j+k}\\
\sfrac{1}{n} & \Cdots & \sfrac{1}{n}& \sfrac{1}{n} & \sfrac{1}{n} & \Cdots & \sfrac{1}{n} & \sfrac{1}{n} & \Cdots & \sfrac{1}{n}\\
 \Vdots & & \Vdots & \Vdots & \Vdots & & \Vdots  &\Vdots & & \Vdots \\
\sfrac{1}{n} & \Cdots & \sfrac{1}{n}& \sfrac{1}{n} & \sfrac{1}{n} & \Cdots & \sfrac{1}{n} & \sfrac{1}{n} & \Cdots & \sfrac{1}{n}\\
\end{pNiceMatrix}}
\end{equation}

where the subscript $B$ stands for \textit{Block}, while the superscripts $(i,j;k)$ remark the dependence on both the row index $j$ and the column index $i$ and the lenght of the block $k$. The corresponding output matrix is
\begin{equation}
S_{B}^{(i,j;k)} V=\begin{pNiceMatrix}[last-col=2]
\mathbb{E}\left[(v^1)^T, \cdots, (v^n)^T\right]\\
\mathbb{E}\left[(v^1)^T, \cdots, (v^n)^T\right]\\
\Vdots \\
\mathbb{E}\left[(v^1)^T, \cdots, (v^n)^T\right]\\
\mathbb{E}\left[(v^i)^T, \cdots, (v^{i+k})^T\right] &\matheqbox{a}{\leftarrow j}\\
\mathbb{E}\left[(v^i)^T, \cdots, (v^{i+k})^T\right]\\
\Vdots \\
\mathbb{E}\left[(v^i)^T, \cdots, (v^{i+k})^T\right] &\matheqbox{a}{\leftarrow j+k}\\
\mathbb{E}\left[(v^1)^T, \cdots, (v^n)^T\right]\\
\mathbb{E}\left[(v^1)^T, \cdots, (v^n)^T\right]\\
\Vdots \\
\mathbb{E}\left[(v^1)^T, \cdots, (v^n)^T\right]\\
\end{pNiceMatrix}
\end{equation}
Notice that only the rows between $j$ and $j+k$ carry on processed information, which are the mean over the interested token values. This observation identifies a highly self-interacting cluster. Lastly, it is worth to remark that it is possible to interpret this situation as a coexistence of small vertical lines. 

\begin{figure}[h!]
    \centering
    {\includegraphics[width=0.8\linewidth]{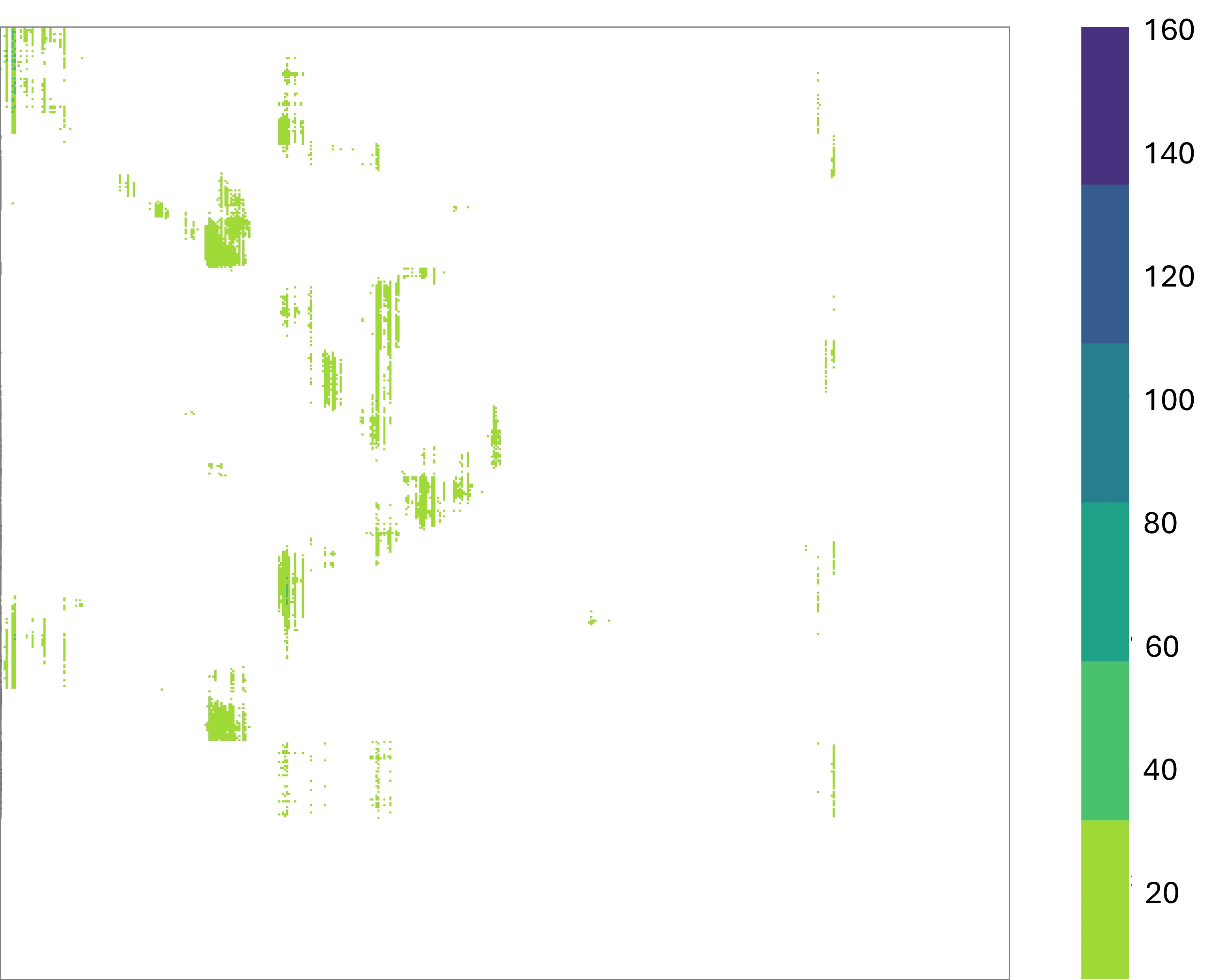}}
    \caption{Attention map of head 1, layer 2 of a correctly classified sequence, showing cluster patterns.}
    \label{fig:omicron_2_1_clu}
\end{figure}

In Fig. \ref{fig:omicron_2_1_clu}, an example of a pattern showing a certain number of cluster patterns is reported. Consider that, in our case study results, they are usually paired with other conformations.

\subsection{Instance gate}
Consider now the case of the self-attention matrix showing a \textbf{pointwise pattern}. It is modelled as
\begin{equation}
S_P^{(i,j)}=\begin{pNiceMatrix}[
  first-row, last-col=8
]
 &   &   &{\begin{array}{c} i \\ \downarrow \end{array}} & & &  \\
\sfrac{1}{n} & \Cdots & \sfrac{1}{n}& \sfrac{1}{n} & \sfrac{1}{n} & \Cdots & \sfrac{1}{n}\\
\Vdots & &  \Vdots & \Vdots & \Vdots &  & \Vdots\\
\sfrac{1}{n} & \Cdots & \sfrac{1}{n}& \sfrac{1}{n} & \sfrac{1}{n} & \Cdots & \sfrac{1}{n}\\
 0 & \Cdots & 0 & \mathbf{1} & 0 & \Cdots & 0 & \matheqbox{a}{\leftarrow j}\\
\sfrac{1}{n} & \Cdots & \sfrac{1}{n}& \sfrac{1}{n} & \sfrac{1}{n} & \Cdots & \sfrac{1}{n}\\
\Vdots & &  \Vdots & \Vdots & \Vdots &  & \Vdots\\
\sfrac{1}{n} & \Cdots & \sfrac{1}{n}& \sfrac{1}{n} & \sfrac{1}{n} & \Cdots & \sfrac{1}{n}\\
\end{pNiceMatrix}
\end{equation}
where the subscript $P$ stands for \textit{Pointwise}, while the superscripts $(i,j)$ remark the dependence on the row index $j$ and the column index $i$. The corresponding output matrix is
\begin{equation}
S_{P}^{(i,j)} V=\begin{pNiceMatrix}[last-col=2]
\mathbb{E}\left[(v^1)^T, \cdots, (v^n)^T\right]\\
\mathbb{E}\left[(v^1)^T, \cdots, (v^n)^T\right]\\
\Vdots \\
\mathbb{E}\left[(v^1)^T, \cdots, (v^n)^T\right]\\
(v^i)^T & \matheqbox{a}{\leftarrow j}\\
\mathbb{E}\left[(v^1)^T, \cdots, (v^n)^T\right]\\
\mathbb{E}\left[(v^1)^T, \cdots, (v^n)^T\right]\\
\Vdots \\
\mathbb{E}\left[(v^1)^T, \cdots, (v^n)^T\right]\\
\end{pNiceMatrix}
\end{equation}
Except for the $j^{th}$ row, the matrix is filled by the characteristical background noise. The information is pointwise and is represented in the output by both the location ($i$) and the components ($j$) of the row, highlighting just one searching direction. The real discrimination role is played by the intensity of the instance: it is indeed usual that sparse points occur in the self-attention matrix, whose low value makes them less significant. 

\subsection{Closed gate}
Finally, consider the case of the self-attention matrix not showing any pattern, i.e. the \textbf{maximal entropy pattern}. It is modelled as
\begin{equation}
S_{ME}=\begin{pNiceMatrix}
\mathbf{\sfrac{1}{n}}  & \mathbf{\sfrac{1}{n}} & \mathbf{\sfrac{1}{n}} & \Cdots & \mathbf{\sfrac{1}{n}} \\
 \mathbf{\sfrac{1}{n}} & \mathbf{\sfrac{1}{n}} & \mathbf{\sfrac{1}{n}} & \Cdots & \mathbf{\sfrac{1}{n}} \\
\mathbf{\sfrac{1}{n}} & \mathbf{\sfrac{1}{n}} & \mathbf{\sfrac{1}{n}}  & \Cdots & \mathbf{\sfrac{1}{n}}\\
\Vdots & \Vdots & \Vdots & \Ddots  & \Vdots \\
\mathbf{\sfrac{1}{n}} & \mathbf{\sfrac{1}{n}} & \mathbf{\sfrac{1}{n}} & \Cdots & \mathbf{\sfrac{1}{n}} \\
\end{pNiceMatrix}
\end{equation}
where the subscript $ME$ stands for \textit{Maximal Entropy}. The corresponding output matrix is
\begin{equation}
S_{P}^{(i,j)} V=\begin{pNiceMatrix}
\mathbb{E}\left[(v^1)^T, \cdots, (v^n)^T\right]\\
\mathbb{E}\left[(v^1)^T, \cdots, (v^n)^T\right]\\
\Vdots \\
\mathbb{E}\left[(v^1)^T, \cdots, (v^n)^T\right]\\
\end{pNiceMatrix}
\end{equation}
which is the configuration of maximal entropy and no processed information conveyed. It highlights that the searching direction cone has pointed to a totally irrelevant subspace which gives no useful information to add to the searching engine process. Here, a rank collapse carrying on mere background noise occurs. That is the reason why it is called \textbf{closed gate}, since the head plays no helpful role and the information is brought forward by the skip connection. 

\subsection{Pattern composition}

Thanks to the linearity of the output computation, the overlap and co-existence of different patterns do not affect the model analysis. As a matter of fact, it is enough to decompose the score matrix into pure patterns and then consider them individually. Nevertheless, this is basically the most common case. In Fig. \ref{fig:omicron_6_2_overlap}, an example of pattern composition is reported. It shows an open gate, backward and forward contextual gates, instance gates, clusters and little directional gates:
\begin{figure}[h!]
    \centering
    {\includegraphics[width=0.8\linewidth]{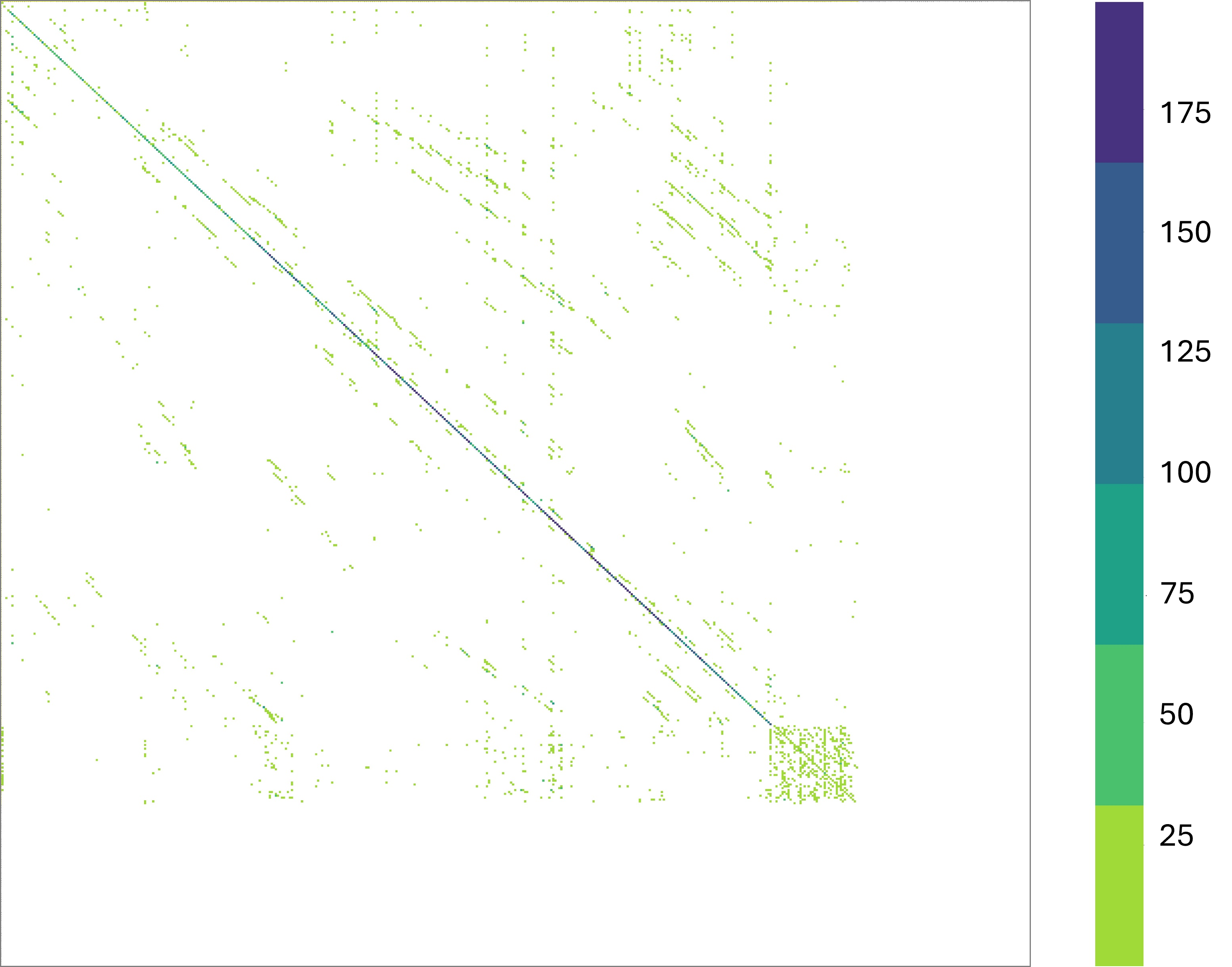}}
    \caption{Attention map of head 2, layer 6 of a correctly classified sequence, showing the co-existence of different patterns.}
    \label{fig:omicron_6_2_overlap}
\end{figure}

\subsection{Inverse directional gate}

It is worthwhile, for the sake of completeness, to mention one last possible pattern, consisting of a \textbf{horizontal line}. Considering that the softmax normalizes the matrix row-wise, and, in a certain way breaks the symmetry between rows and columns, it is modelled as
\begin{equation}
S_H^{(j)}=\begin{pNiceMatrix}[
   last-col=8
]
\sfrac{1}{n} & \Cdots & \sfrac{1}{n}& \sfrac{1}{n} & \sfrac{1}{n} & \Cdots & \sfrac{1}{n}\\
\Vdots & &  \Vdots & \Vdots & \Vdots &  & \Vdots\\
\sfrac{1}{n} & \Cdots & \sfrac{1}{n}& \sfrac{1}{n} & \sfrac{1}{n} & \Cdots & \sfrac{1}{n}\\
 \mathbf{\ast} & \Cdots &  \mathbf{\ast} &  \mathbf{\ast} &  \mathbf{\ast} & \Cdots &  \mathbf{\ast} & \matheqbox{a}{\leftarrow j}\\
\sfrac{1}{n} & \Cdots & \sfrac{1}{n}& \sfrac{1}{n} & \sfrac{1}{n} & \Cdots & \sfrac{1}{n}\\
\Vdots & &  \Vdots & \Vdots & \Vdots &  & \Vdots\\
\sfrac{1}{n} & \Cdots & \sfrac{1}{n}& \sfrac{1}{n} & \sfrac{1}{n} & \Cdots & \sfrac{1}{n}\\
\end{pNiceMatrix}
\end{equation}
where the subscript $H$ stands for \textit{Horizontal}, while the superscript $(j)$ remarks the dependence on the row index $j$. Here, for the pattern to be significant, it is needed that only a small number of entries in the $j^{th}$ row are different from 0: this set is called $\mathcal{K}$. Therefore, the corresponding output matrix is
\begin{equation}
S_{P}^{(i,j)} V=\begin{pNiceMatrix}[last-col=2]
\mathbb{E}\left[(v^1)^T, \cdots, (v^n)^T\right]\\
\mathbb{E}\left[(v^1)^T, \cdots, (v^n)^T\right]\\
\Vdots \\
\mathbb{E}\left[(v^1)^T, \cdots, (v^n)^T\right]\\
\mathbb{E}\left[\{(v^i)^T\}_{i \in \mathcal{K}}\right] & \matheqbox{a}{\leftarrow j}\\
\mathbb{E}\left[(v^1)^T, \cdots, (v^n)^T\right]\\
\mathbb{E}\left[(v^1)^T, \cdots, (v^n)^T\right]\\
\Vdots \\
\mathbb{E}\left[(v^1)^T, \cdots, (v^n)^T\right]\\
\end{pNiceMatrix}
\end{equation}
The denomination \textbf{inverse directional gate} comes from the pattern being in some way specular to the directional one. This configuration is quite delicate, since, when it occurs, it consists of so few points (whose order of magnitude is some units) that it seems almost better to reinterpret it as a composition of instances.

\section{Global analysis}

\subsection{Cone index results on spectral analysis}

\begin{figure*}[h!]
    \centering
    {\includegraphics[width=\linewidth]{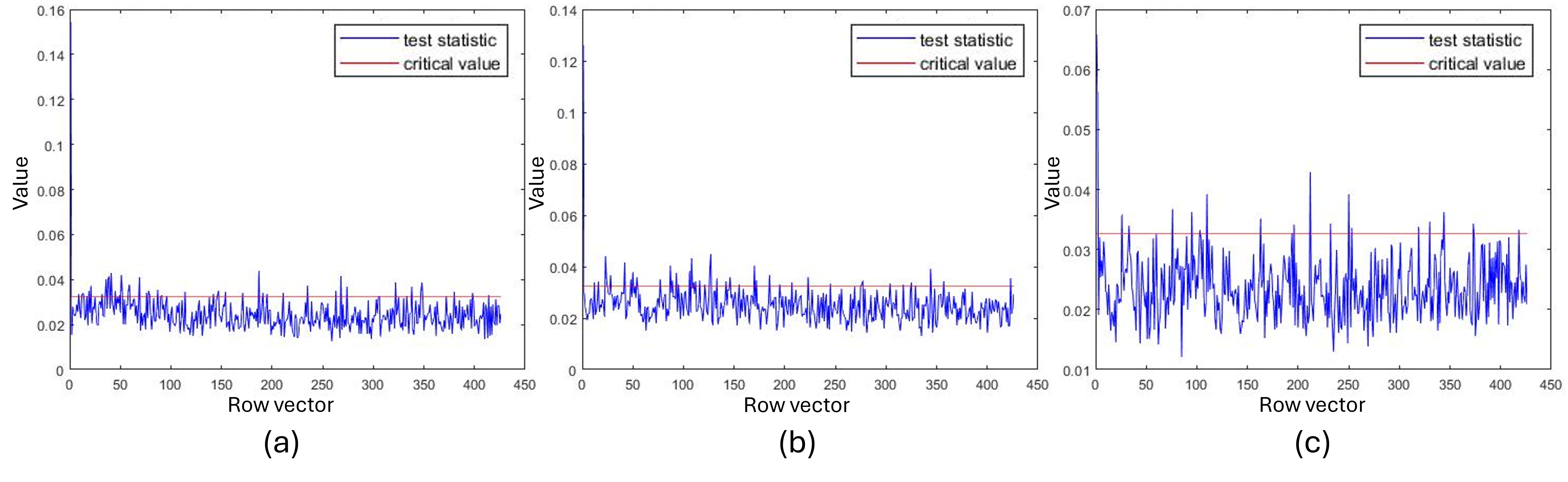}}
    \caption{Normality Lillefors test after the first LN, for layers 2 (a), 3 (b) and 6 (c).}
    \label{fig:ll2_ll3_ll6}
\end{figure*}

This last section is devoted to discussing the overall behaviour of the processed data throughout the entire Transformer. The first results concerns the final distribution of the coordinates of the examined row-vectors which turn out to be Gaussian distributed. Since the LN does not affect the distribution, as it is just the composition of a translation with a rescaling, it has been possible to run the normality Lillefors test after the first LN, for each layer. Remarking that the test critical statistic value for rejecting the normality hypothesis is $k=0,032$ (identified by the red line), Fig. \ref{fig:ll2_ll3_ll6} reports the resulting statistic for layers 2, 3 and 6.

\noindent Notice that there are only a few rows which fail the test, while the great majority really corresponds to Gaussian vectors. The next step of the analysis is the study of the coherence between these vectors. The technique adopted consists of summing all the output row vectors and analyzing the resulting feature distribution. It turns out that not only correlation exists since the distribution is not uniform, but also the vector is Gaussian shaped. In Fig. \ref{fig:ss2_ss3_ss6}, the corresponding histograms of layers 2, 3 and 6 are reported:
\begin{figure*}[h!]
    \centering
    {\includegraphics[width=\linewidth]{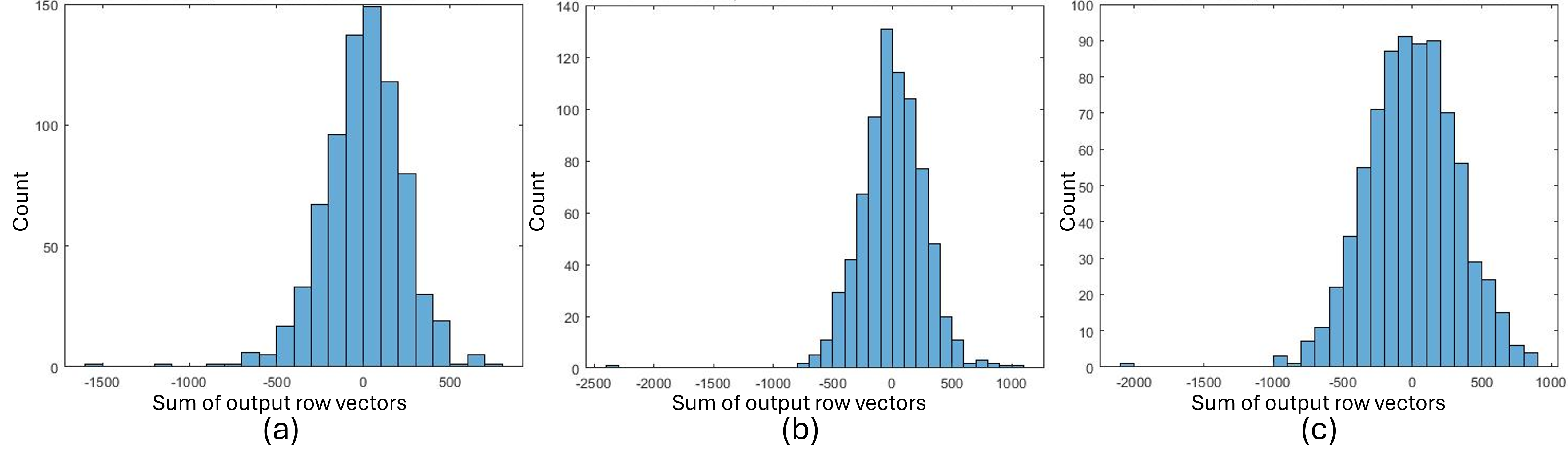}}
    \caption{Histograms of the sum of all the output row vectors of layer 2 (a), 3 (b) and 6 (c), showing the mean distribution of the features.}
    \label{fig:ss2_ss3_ss6}
\end{figure*}

Additionally, Table \ref{tab:lillefors} reports the statistics of the Lillefors test. Even if it is possible to recognize a normally shaped distribution, the first 4 layers fail the test while the last 8 layers pass it, highlighting the normal shaping process.

\begin{table}[H]
    \centering
    \caption{Normality Lillefors test for the output in each layer.}
    \begin{tabular}{|l|l||l|l||l|l|}
    \hline
    Layer & Statistic &Layer & Statistic & Layer & Statistic\\
    \hline
    1 & 0,060 & 5 & 0,017 & 9 & 0,019 \\
    \hline
    2 & 0,052 & 6 & 0,015 & 10 & 0,014\\
    \hline
    3 & 0,033 & 7 &0,026 & 11 & 0,019 \\
    \hline
    4 & 0,040 & 8 &0,017 & 12 & 0,015 \\
    \hline
    \end{tabular}
    \label{tab:lillefors}
\end{table}

\noindent The mentioned failure is due to the presence of oversized extremal bins, which highlights the deep directional selection, more relevant in the initial layers where the network has to find the right subspaces.

\par

\begin{figure}[h!]
\centering
{\includegraphics[width=0.7\linewidth]{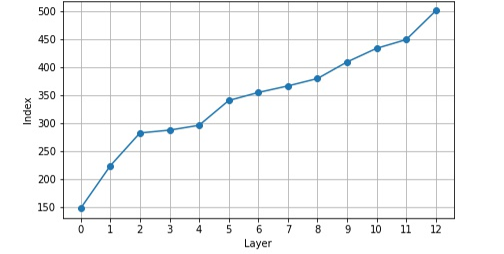}}
\caption{Cone index of a random sequence correctly classified.}
\label{fig:con_ind}
\end{figure}

\begin{figure*}[h!]
    \centering
    {\includegraphics[width=\linewidth]{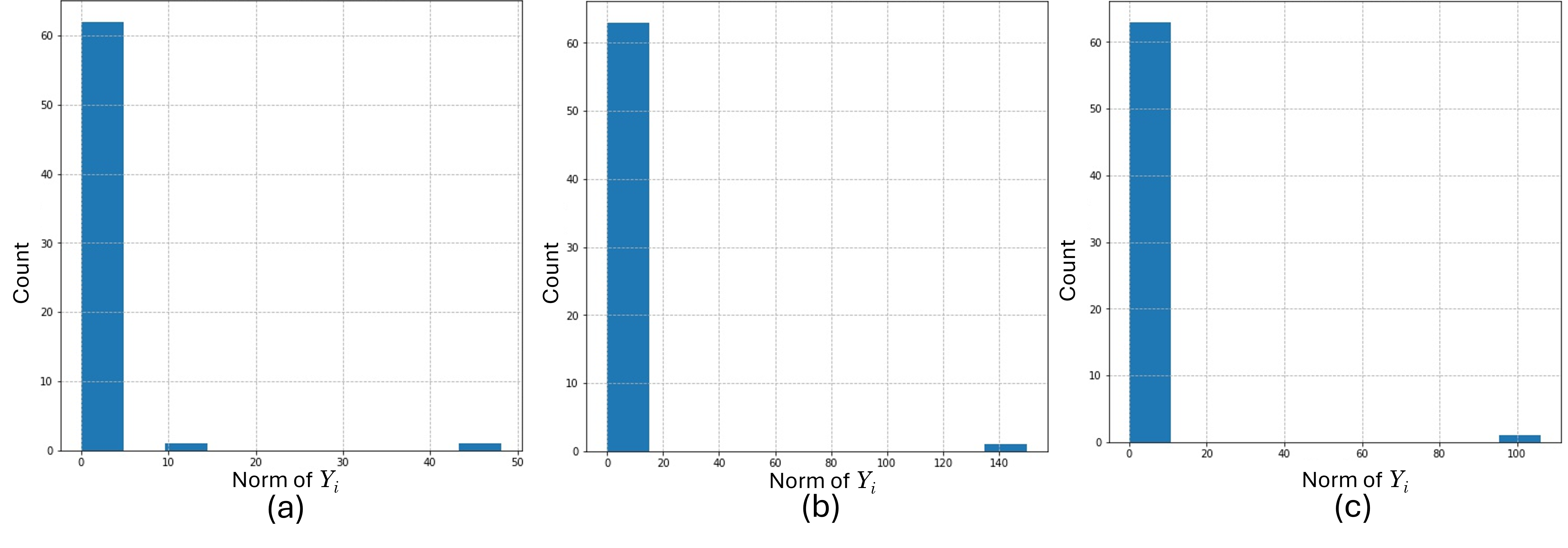}}
    \caption{Histograms of singular values of $Y_i$ output for head 4 of layer 1 (a), head 10 of layer 5 (b), and head 9 in layer 12 (c).}
    \label{fig:sing_val_Yi_1_4_Yi_5_10_Yi_12_9}
\end{figure*}

The coherence between the vectors is an extremely indicative notion in order to understand the Transformer neural network. The propagation model presented in the paper forecasts a global vector alignment behaviour, due to the network's purpose to find relevant directions. Therefore, the  so-called \textbf{cone index} is introduced, in order to quantify the coherence:
\begin{equation}
I_{cone}([l])=\| \sum_{\mathbf{x} \in [l]} \mathbf{x} \|
\end{equation}
where $\mathbf{x}$ denotes the generic row-vector of the resulting matrix after the second layer normalization and $[l]$ makes explicit the dependence on the layer. Fig. \ref{fig:con_ind} shows the cone index of a random correctly classified sequence. According to the model, it is possible to witness layer-increasing behaviour
which shows the Transformer inclination to select specific subspaces. A further, although different, corroboration consists in checking the singular values of the output matrices. Fig. \ref{fig:sing_val_Yi_1_4_Yi_5_10_Yi_12_9}(a)-(c) shows the histograms of singular values of $Y_i$ output for head 4 of layer 1, head 10 of layer 5, and head 9 in layer 12, respectively. As expected, the maximal singular value tends to increase with the layers, selecting less and less preferential directions.

\subsection{Information content}

\begin{figure}[h!]
\centering
{\includegraphics[width=0.6\linewidth]{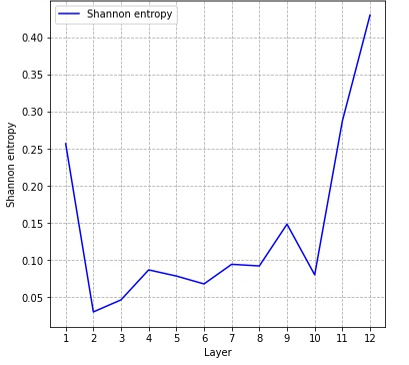}}
\caption{Layer-wise Shannon entropy plot.}
\label{fig:Shannon_entropy_plot}
\end{figure}

The Shannon entropy is employed as an informative indicator in the $S$ matrices (see e.g. \cite{entr}) in order to quantify the captured information content. Larger entropy values highlight the absence of patterns, which is related to minor information content, while smaller values identify captured information. In Fig. \ref{fig:Shannon_entropy_plot}, the layer-wise Shannon entropy plot is reported. 
Notice that the lower layers have specialized in finding the information content, that is the direction choices, while the upper ones seem just to bring forward the previous achievements.

\par 

\begin{figure}[h!]
    \centering
    {\includegraphics[width=\linewidth]{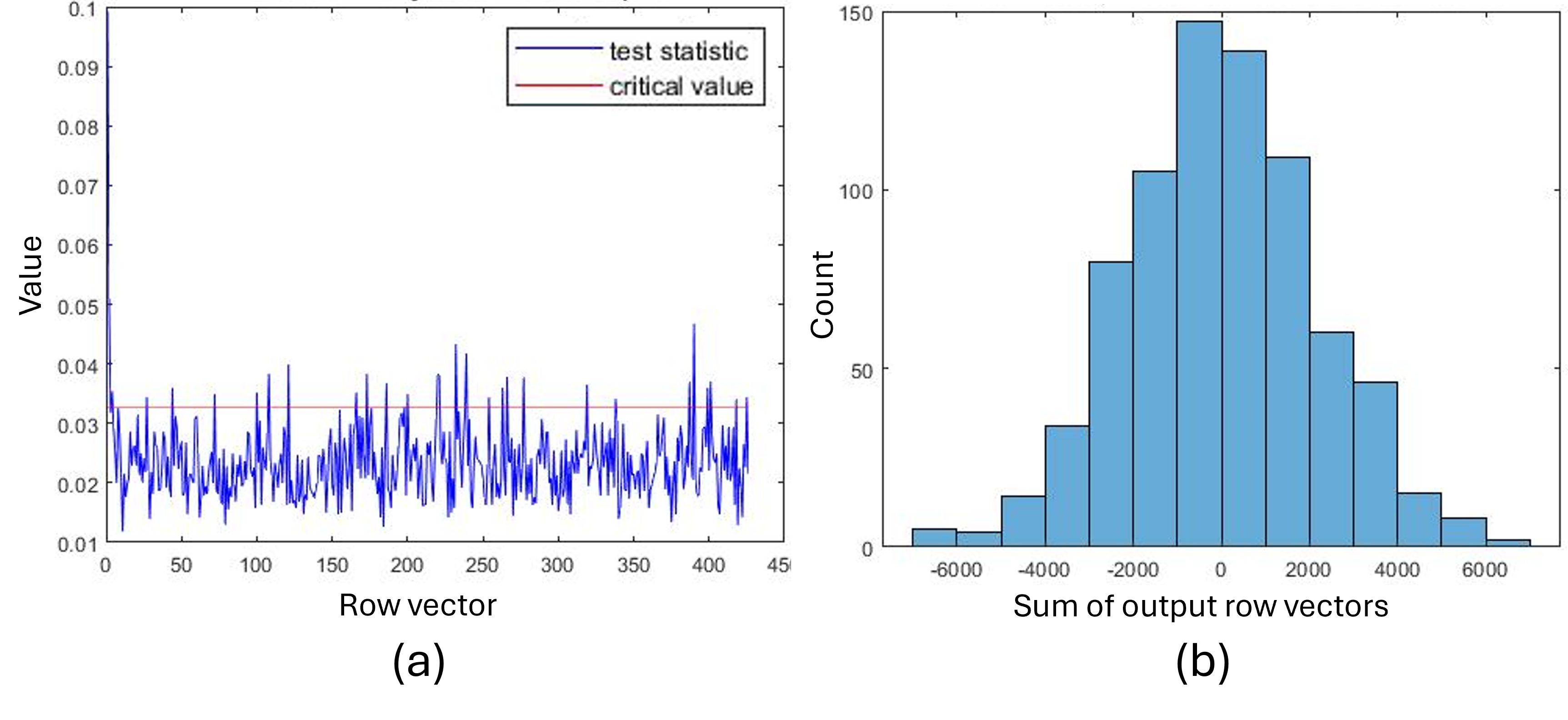}}
    \caption{Histogram and Lillefors plot of the sum of all the output vectors in all the layers.}
    \label{fig:llall_ssall}
\end{figure}

Finally, in order to give a global idea about the location of the relevant information, the distribution of the vector consisting of the sum of all the output vectors in all the layers has been checked. Quite surprisingly, it provided a normally distributed sample centered at zero, whose histogram and Lillefors plot are reported in Fig. \ref{fig:llall_ssall}.

\subsection{Universal analysis}

The aim of this subsection focuses on studying the propagation of the information content along the Transformer. It is shown that the information content stream behaves in a specific way which globally is not channel-wise, despite in some small and limited tracks, in contrast with the immediate intuition arising from the fact that the heads are concatenated in a predetermined order. 

\subsubsection{Technical results}
First, it is worth recalling some general facts and definitions which will be useful in the sequel. Specifically, the concept of the concatenation operator acting column-wise, and its relative transpose acting row-wise, is needed.

\begin{defn}
Let $X^{(i)} \in \mathbb{R}^{r, s}$ be a finite family of matrices labelled by the index $i=1, \dots, n$. Their column-wise concatenation 
\begin{equation}
\mathrm{Conc}_{i=1}^n \left(X^{(i)} \right)=X
\end{equation}
corresponds to a matrix $X \in \mathbb{R}^{r , s \cdot n}$ such that the generic element $x_{j,k}$ is given by
\begin{equation}
x_{j,k}=x^{(i)}_{j,k \, \mathrm{mod} \, s} \in X^{(i)} \text{ where } i=\frac{ k -(k \, \mathrm{mod} \, s) +1}{n}.
\end{equation}
\end{defn}

\begin{rem}
Intuitively the operator $\mathrm{Conc}_{i=1}^n$ corresponds to a \textit{horizontal} concatenation.
\end{rem}
The next result can be proved by straightforward computations. 
\begin{prop}
Let $X^{(i)} \in \mathbb{R}^{r , s}$ be a finite family of matrices labelled by the index $i=1, \dots, n$, and let $Y \in \mathbb{R}^{p , r}$. Then:
\begin{equation}
Y \cdot \mathrm{Conc}_{i=1}^n \left(X^{(i)} \right)=\mathrm{Conc}_{i=1}^n \left(Y \cdot X^{(i)} \right)
\end{equation}
\end{prop}

\begin{defn}
Let $X^{(i)} \in \mathbb{R}^{r, s}$ be a finite family of matrices labelled by the index $i=1, \dots, n$. Their row-wise concatenation 
\begin{equation}
{}^{T}\mathrm{Conc}_{i=1}^n \left(X^{(i)} \right)=X
\end{equation}
corresponds to a matrix $X \in \mathbb{R}^{r \cdot n , s }$ such that the generic element $x_{j,k}$ is given by:
\begin{equation}
x_{j,k}=x^{(i)}_{j \, \mathrm{mod} \, r,k } \in X^{(i)} \text{ where } i=\frac{ j -(j \, \mathrm{mod} \, r) +1}{n}.
\end{equation}
\end{defn}

\begin{rem}
As one can expect, the operator ${}^{T}\mathrm{Conc}_{i=1}^n$ intuitively corresponds to a \textit{vertical} concatenation.
\end{rem}

\begin{prop}\label{propimp}
For any family of matrices $X^{(i)} \in \mathbb{R}^{r , p}$ and $Y^{(i)} \in \mathbb{R}^{p , s }$, $i=1, \dots, n$ the following equality holds true:
\begin{equation}
 \mathrm{Conc}_{i=1}^n \left(X^{(i)} \right) \cdot  {}^T\mathrm{Conc}_{i=1}^n \left(Y^{(i)} \right) =\sum_{i=1}^n \left(X^{(i)} \cdot Y^{(i)} \right) \in \mathbb{R}^{r , s}
\end{equation}
\end{prop}

In order to prove the claim is sufficient to notice that the result of the concatenation operators are block matrices. Then, the result follows by direct computations.

\subsubsection{Multi-head attention}

The interest of such analysis focuses on modelling the information diffusion along the Transformer levels. To this aim, it is convenient to analyze the information stream from the multi-head attention block, after the searching engine process, where the information content arises.

Notice that, as a result of the self-attention process, every head yields an output matrix of dimension $512 \times 64$, which are consequently concatenated. Formally, labelling each head with the index $i, i=1, \dots, 12$ and denoting the resulting outputs by 
$Y^{(i)}=S \cdot V^{(i)}$, one has that the informative content is represented by the matrix 
\begin{equation}
Y_{MH}=\mathrm{Conc}_{i=1}^n \left(Y^{(i)} \right)
\end{equation}
where the subscript MH refers to the Multi-Head level. Notice that $Y_{MH} \in \mathbb{R}^{512,768}$.

\subsubsection{Skip connection and layer normalization}

The skip connection 
\begin{equation}
\begin{aligned}
\mathrm{LN}(Y_{MH}+X)=&\mathrm{LN}\left(\mathrm{Conc}_{i=1}^n \left(Y^{(i)} + X^{(i)} \right)\right)\\
=&D^{-1}\left\{ \mathrm{Conc}_{i=1}^n \left(Y^{(i)} + X^{(i)} \right)\right. \\
& - \left. \mathbf{1} \mathbb{E}\left(\mathrm{Conc}_{i=1}^n \left(Y^{(i)} + X^{(i)} \right)\right)^T \right\}
\end{aligned}
\end{equation}
can be viewed as the linear transformation which performs a sort of subspace reset within the searching engine process. This mechanism basically consists of projecting the information conveyed by the matrices $\mathrm{Conc}_{i=1}^n \left(Y^{(i)} + X^{(i)} \right)$ in row-wise probability spaces where the mutual influence between the features plays a sort of rescaling role. Even if it slightly affects the numerical values, it does not add any real contextual information, since the transformation is related to the architecture of the Transformer rather than the external inputs. With respect to this process, it is reasonable to remark that the result of the skip connection and layer normalization in the information streaming study may be rewritten as
\begin{equation}
\mathrm{LN}\left(\mathrm{Conc}_{i=1}^n \left(Y^{(i)} + X^{(i)} \right)\right) = \mathrm{Conc}_{i=1}^n \widetilde{Y}^{(i)}
\end{equation}
where the information conveyd by each $\widetilde{Y}^{(i)}$ depends only on $Y^{(i)}$ and $X^{(i)}$. This means that the skip connection and layer normalization process just acts as a channel from the information streaming point of view.

\subsubsection{Multi-layer perceptron}

The multi-layer perceptron takes the skip connection and layer normalization output and acts row-wise in this way:
\begin{equation}
\begin{aligned}
&\mathrm{max}\left\{ \left( \mathrm{Conc}_{i=1}^n \widetilde{Y}^{(i)} \right) W_1 + \mathbf{1}(b_1)^T, 0 \right\} W_2 + \mathbf{1}(b_2)^T
\end{aligned}
\end{equation}
Splitting the matrix $W_1$ into horizontal blocks by means of the operator ${}^T \mathrm{Conc}$, the previous expression can be rewritten in a more convenient way as:
\begin{equation}
\mathrm{max}\left\{ \left( \mathrm{Conc}_{i=1}^n \widetilde{Y}^{(i)} \right) \cdot \left( {}^T \mathrm{Conc}_{i=1}^n W_1^{(i)} \right) + \mathbf{1}(b_1)^T, 0 \right\} W_2 +  \mathbf{1}(b_2)^T
\end{equation}
Hence, employing Proposition $\ref{propimp}$, one obtains that:
\begin{equation}
\mathrm{max}\left\{ \sum_{i=1}^n \left(\widetilde{Y}^{(i)} \cdot W_1^{(i)} \right) + \mathbf{1}(b_1)^T, 0 \right\} W_2 +  \mathbf{1}(b_2)^T
\end{equation}
Consider now the $\mathrm{max}$ operator: it just annihilates the eventual negative values. In the information content stream study, it does not interact between features, since it only can possibly select them. For this reason, the latter expression can be rewritten in terms of new elements $\widetilde{Y'}^{(i)}$ depending only on $\widetilde{Y}^{(i)}$ with eventual 0 where the ReLU behaves:
\begin{equation}
\begin{aligned}
&\left[ \sum_{i=1}^n \left(\widetilde{Y'}^{(i)} \cdot W_1^{(i)} \right) + \mathbf{1}(b_1)^T \right] W_2 +  \mathbf{1}(b_2)^T\\
=&\left(\sum_{i=1}^n \widetilde{Y'}^{(i)} \cdot W_1^{(i)} \cdot W_2 \right) +\left( \mathbf{1}(b_1)^T \cdot W_2 + \mathbf{1}(b_2)^T\right)\\
=&\left(\sum_{i=1}^n \widetilde{Y'}^{(i)} \cdot \widetilde{W}^{(i)} \right) +\left( \mathbf{1}(b_1)^T \cdot W_2 + \mathbf{1}(b_2)^T\right)
\end{aligned}
\end{equation}
where the terms $\widetilde{W}^{(i)}$ just represent the multiplication between $W_1^{(i)}$ and $W_2$. In the last expression, it is possible to recognize 3 different terms: the first one depending on the input and conveying the information content, and a bias term, actually consisting of two different terms, a uniform bias $(\mathbf{1}(b_2)^T)$ and an anisotropic bias $(\mathbf{1}(b_1)^T \cdot W_2)$.

Notice that, as a result of the multi-layer perceptron block, the order information regarding the head position is completely lost, since the concatenation operator has given way to the sum. It is worth to remark that the multi-layer perceptron performs the crucial role of synthesizing the information content by means of a sort of order delete which makes the subsequent searching engine mechanism unbiased. 

Finally, it acts non-linearly and adds anisotropy in the network economy, while the uniform bias seems redundant.

\subsubsection{Searching engine process}

Each head in the next level receives the same input which is a synthesis of the information present in the previous level. It does not arise any subspace specialization, since the information deriving from the different heads is summed up. It is the \textbf{mean information content} which is conveyed to the next level and processed.

As already remarked before, the skip connection and layer normalization mechanism do not substantially affect the information content, therefore the same notation with respect to the multi-layer perceptron output is employed.

Recalling the searching engine process, one has that
\begin{equation}
\begin{aligned}
QK^T & =(X W_Q +\mathbf{1} b_Q^T)(X W_K + \mathbf{1} b_K^T)^T\\
&=X W_Q W_K^T X^T  + \mathbf{1}b_Q^T W_K^T X^T + X W_Q b_K \mathbf{1}^T+ \mathbf{1} b_Q^T b_K \mathbf{1}^T\\
&=X M X^T + \mathbf{1}b_Q^T W_K^T X^T + X W_Q b_K \mathbf{1}^T+ \mathbf{1} b_Q^T b_K \mathbf{1}^T
\end{aligned}
\end{equation}
and now, substituting $X$ with the level input $\left(\sum_{i=1}^n \widetilde{Y'}^{(i)} \cdot \widetilde{W}^{(i)} \right) +\left( \mathbf{1}(b_1)^T \cdot W_2 + \mathbf{1}(b_2)^T\right)$, the equation becomes:
\begin{equation}
\begin{aligned}
QK^T  =&\left( \left(\sum_{i=1}^n \widetilde{Y'}^{(i)} \cdot \widetilde{W}^{(i)} \right) +\left( \mathbf{1}(b_1)^T \cdot W_2 + \mathbf{1}(b_2)^T\right) \right) M \cdot \\
\cdot & \left( \left(\sum_{i=1}^n \widetilde{Y'}^{(i)} \cdot \widetilde{W}^{(i)} \right) +\left( \mathbf{1}(b_1)^T \cdot W_2 + \mathbf{1}(b_2)^T\right) \right)^T +
\\+& \mathbf{1}b_Q^T W_K^T \left( \left(\sum_{i=1}^n \widetilde{Y'}^{(i)} \cdot \widetilde{W}^{(i)} \right) +\left( \mathbf{1}(b_1)^T \cdot W_2 + \mathbf{1}(b_2)^T\right) \right)^T + \\
+& \left( \left(\sum_{i=1}^n \widetilde{Y'}^{(i)} \cdot \widetilde{W}^{(i)} \right) +\left( \mathbf{1}(b_1)^T \cdot W_2 + \mathbf{1}(b_2)^T\right) \right) W_Q b_K \mathbf{1}^T+\\
+& \mathbf{1} b_Q^T b_K \mathbf{1}^T
\end{aligned}
\end{equation}
where it is possible to recognize the contextual information term given by
\begin{equation}
\begin{aligned}
&\left(\sum_{i=1}^n \widetilde{Y'}^{(i)} \widetilde{W}^{(i)} \right) M \left(\sum_{i=1}^n \widetilde{Y'}^{(i)} \widetilde{W}^{(i)} \right)\\
=& \sum_{i=1}^n \sum_{j=1}^n  \widetilde{Y'}^{(i)} \widetilde{W}^{(i)}  M \widetilde{Y'}^{(j)} \widetilde{W}^{(j)} 
\end{aligned}
\end{equation}
and added to different anisotropic and uniform bias terms.

\subsubsection{Final comments}

It has been shown that the information content in BERT globally propagates between two layers by conveying a mean information content. The channel structure is totally broken by the behaviour of the multi-layer perceptron, which actually synthesizes the information blocks received in input. The subsequent searching engine process is then free to behave unbiasedly selecting the convenient subspaces on the basis of the mere input and not of an architectural artifice.

\section{Conclusions and future work}
The increasing importance of explainability and interpretability in Transformer models, particularly BERT, has motivated efforts to uncover the mathematical mechanisms behind its classification process.
This paper proposes new perspectives on BERT inference mechanisms, whose attention-seeking consists of a subspace selection driven by the searching engine process. The analysis concerns both the local and the global behaviour of the network. At the level of the single heads, the paper introduces a model for the similarity measure in the attention mechanism, as well as a full study of possible patterns arising in the self-attention matrix. Finally, the semantic content at the level of the information stream is studied by means of the distribution analysis of the data and global statistics as the cone index and Shannon entropy.
Future work could involve the analysis of the BERT learning process during training, examining the specialization of individual components in the BERT architecture and leveraging these findings to optimize the architecture of the Transformer encoder.

\section{Acknowledgements}
Funding: This publication is part of the project PNRR-NGEU which has received funding from the MUR – DM 352/2022.

 \bibliographystyle{elsarticle-num} 
 \bibliography{references.bib}

\end{document}